\pdfoutput=1

\documentclass[11pt]{article}

\usepackage{acl}
\usepackage{times}
\usepackage{latexsym}
\usepackage{graphicx}

\usepackage[T1]{fontenc}
\usepackage[utf8]{inputenc}

\usepackage{microtype}
\usepackage{booktabs}
\usepackage{placeins}
\usepackage{wrapfig}

\usepackage{amsmath}
\usepackage{amssymb}

\usepackage{inconsolata}

\usepackage{subcaption}

\title{Overconfidence is Key: Verbalized Uncertainty Evaluation in Large Language and Vision-Language Models}

\author{Tobias Groot \quad Matias Valdenegro-Toro\\
  Department of Artificial Intelligence, University of Groningen.\\
  \texttt{m.a.valdenegro.toro@rug.nl}}

\begin{document}
\maketitle
\begin{abstract}
Language and Vision-Language Models (LLMs/VLMs) have revolutionized the field of AI by their ability to generate human-like text and understand images, but ensuring their reliability is crucial. This paper aims to evaluate the ability of LLMs (GPT4, GPT-3.5, LLaMA2, and PaLM 2) and VLMs (GPT4V and Gemini Pro Vision) to estimate their verbalized uncertainty via prompting. We propose the new Japanese Uncertain Scenes (JUS) dataset, aimed at testing VLM capabilities via difficult queries and object counting, and the Net Calibration Error (NCE) to measure direction of miscalibration.
Results show that both LLMs and VLMs have a high calibration error and are overconfident most of the time, indicating a poor capability for uncertainty estimation. Additionally we develop prompts for regression tasks, and we show that VLMs have poor calibration when producing mean/standard deviation and 95\% confidence intervals.
\end{abstract}

\section{Introduction}

\begin{figure}[t]
    \centering
    \begin{subfigure}{0.45\linewidth}
        \includegraphics[width=\linewidth]{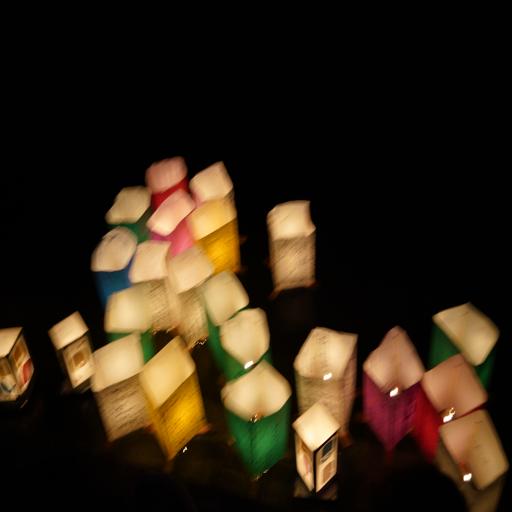}\\
    \end{subfigure}
    \,\,
    \begin{subfigure}{0.50\linewidth}
        \textbf{Prompt}: How many lamps are shown in this photo?  Moreover, please express your estimate as a 95\% confidence interval. Format your answer as:’[Lower Bound, Upper Bound]’\\        
    \end{subfigure}
    \textbf{GPT-4V}: [12, 22]. \textbf{GPV}: [15, 19]. \textbf{Correct}: 23
    \caption{Example prompt results for GPT-4V and Gemini Pro Vision on a JUS Prompt \hyperref[tab:dataset3]{16}, where a 95\% confidence interval is requested but the correct answer is outside the confidence interval. . This shows that VLMs also have problems with verbalized uncertainty, and provide overconfident answers. GPT4-V is closer to the correct answer. Full prompt is provided in Sec \ref{sup:prompt_eng}. Photo taken at the Tōrō-Nagashi on August 6, Hiroshima, Japan (Floating Lantern Ceremony).}
    \label{fig:sushi_prompt}
\end{figure}

Large Language Models (LLMs) and Vision Language Models (VLMs) have been praised for their impressive capabilities across a wide range of applications. However, they are not immune to generating misleading or incorrect information, often referred to as "hallucinations" \cite{huang2023survey}, as illustrated in Figure \ref{fig:sushi_prompt}. This raises a critical question: how can someone know when an answer prompt can be trusted? Usually it is expected that model confidence or probability is a proxy for correctness, incorrect predictions should have low confidence, while correct predictions should have high confidence. Uncertainty estimation has been a valuable tool in assessing the reliability of machine learning models, but the quality of uncertainty estimation within LLMs and VLMs remains relatively underexplored \cite{xiong2023can, huang2023look,kostumov2024uncertainty}. 

This study aims to expand the domain of uncertainty estimation in LLMs by comparing four state-of-the-art LLMs: GPT-3.5, GPT-4 \cite{achiam2023gpt}, LLaMA-2-70b \cite{touvron2023llama}, and PaLM 2 \cite{anil2023palm}, across three distinct NLP tasks: sentiment analysis, math word problems, and named-entity recognition. Additionally, the quality of uncertainty estimation in VLMs is evaluated by testing two newly introduced VLMs, GPT-4V \cite{GPT-4V} and Gemini Pro Vision \cite{team2023gemini}, on a novel image recognition task. 

\begin{figure*}[t!]
    \centering
    \begin{tabular}{ccc}

        \includegraphics[width=0.32\textwidth]{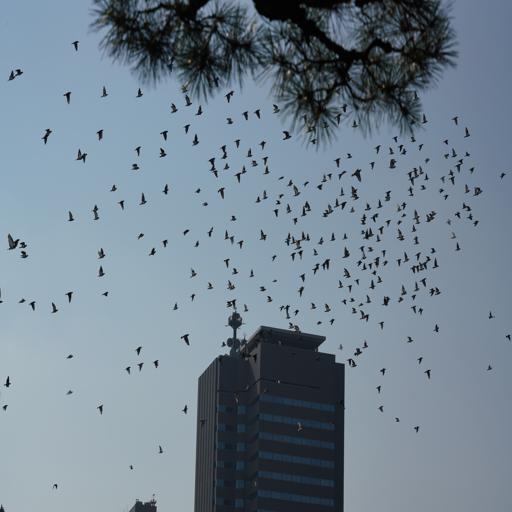} &
        \includegraphics[width=0.32\textwidth]{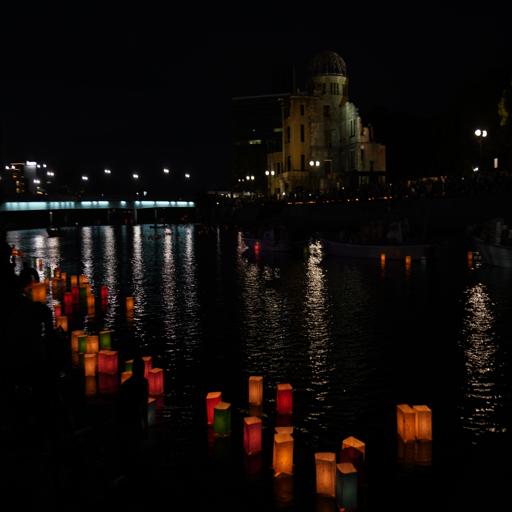} &
        \includegraphics[width=0.32\textwidth, angle=+90]{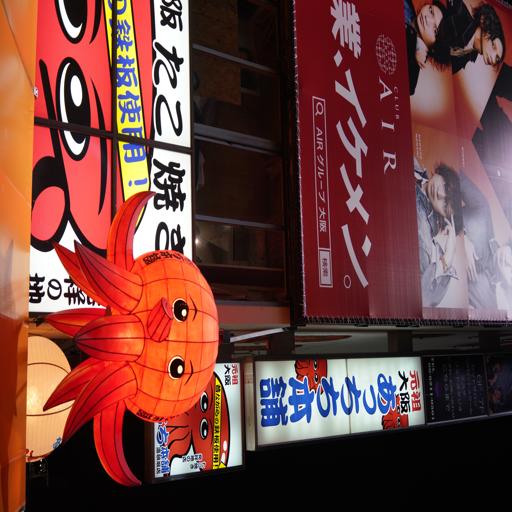} \\
        \parbox{0.32\textwidth}{ \textbf{Prompt: }How many birds are shown in this photo? Moreover, please rate your confidence in your answer between 0 and 100\%. The answer should be in the format: "Answer (confidence\%)".} &
        \parbox{0.32\textwidth}{ \textbf{Prompt:} Where was this photo taken? Moreover, please rate your confidence in your answer between 0 and 100\%. The answer should be in the format: "Answer (confidence\%)".} &
        \parbox{0.32\textwidth}{ \textbf{Prompt:} How many octopuses are shown in this photo? Moreover, please rate your confidence in your answer between 0 and 100\%. The answer should be in the format: "Answer (confidence\%)". } \\
        \multicolumn{2}{c}{} \\
        \parbox[t]{0.32\textwidth}{\textbf{Answer GPT-4V}: It is not feasible to count the exact number of birds in the photo due to their movement, the potential for overlap, and the sheer number (confidence 0\%). The image shows a flock of birds in flight, which makes it particularly challenging to provide an accurate count from a single static image.} &
        \parbox[t]{0.32\textwidth}{\textbf{Answer GPT-4V}: Hiroshima, Japan, during a lantern floating ceremony (confidence 90\%).} &
        \parbox[t]{0.32\textwidth}{\textbf{Answer GPT-4V}: 1 octopus (confidence 100\%)}
        
        \\
        \multicolumn{2}{c}{} \\
        
        \parbox{0.32\textwidth}{\textbf{Answer Gemini Pro Vision}: 100 (80\%)} &
        \parbox{0.32\textwidth}{\textbf{Answer Gemini Pro Vision}: Hiroshima, Japan (80\%)} &
        \parbox{0.32\textwidth}{\textbf{Answer Gemini Pro Vision}: 2 (100\%)} 
        
        \\
        \multicolumn{2}{c}{} \\
        
        \parbox[t]{0.32\textwidth}{\textbf{Correct Answer}: Over 100 birds, hard to count the exact number.} &
        \parbox[t]{0.32\textwidth}{\textbf{Correct Answer}: Japan, Hiroshima, Atomic Bomb Dome } &
        \parbox[t]{0.32\textwidth}{\textbf{Correct Answer}: 5} \\
        
        \multicolumn{2}{c}{} \\
        
    \end{tabular}
    \caption{Example answers from GPT-4V and Gemini Pro Vision for the image recognition task on three JUS image-prompts. Columns 1 and 3 are incorrect overconfident answers, and Column 2 is underconfident correct. These results show how VLMs produce incorrect verbalized uncertainty.}
    \label{tab:vlm_dataset_results}
\end{figure*}

This exploration is guided by the following two research questions: How accurately do Large Language Models estimate their uncertainty across various natural language processing tasks? How accurately do Vision Language Models estimate their uncertainty when performing an image recognition task?

To answer these questions, the level of calibration is evaluated by asking the models to verbalize their confidence alongside their answers. By comparing these confidence levels with their corresponding accuracies, the models’ calibration quality can be assessed.   

The contributions of this paper are: We evaluate VLM and LLM's verbalized uncertainty (Sec \ref{sec:results}). We introduce a novel image recognition dataset, the Japanese Uncertain Scenes, specifically designed for testing the uncertainty estimation capabilities of VLMs via difficult to interpret images and object counting in Sec \ref{sec:jus_dataset}. Furthermore, we propose a new calibration metric, the Net Calibration Error (NCE), which offers insight into the direction of a model's miscalibration in Sec \ref{sec:calib_errors}. We finally evaluate VLM verbalized uncertainty in our proposed dataset, including standard classification percentage confidences, and regression mean/standard deviation and 95\% confidence intervals in Sec \ref{supp:task_mean_std}.

\section{Related Work}
\citet{martino2023exploring} evaluated the uncertainty estimation capabilities of ChatGPT by asking the model to output its confidence in its answer and see if they are well-calibrated. This was done by comparing the accuracy with the outputted confidence in two NLP tasks: sentiment analysis and common sense reasoning. The tasks were performed in five different high-resource languages (English, French, German, Italian, and Spanish) to evaluate if ChatGPT is equally accurate in these languages. The results showed that all languages achieved similar accuracy in both tasks and that ChatGPT is often overconfident and seems to be unaware when it lacks the knowledge to correctly handle an input.

\citet{jiang2021can} researched the calibration of BART, T5, and GPT-2 on question-answering tasks and found that these models are overconfident and thus are not well-calibrated. 

Additionally, \citet{chen2022close} evaluated if pre-trained models (PLMs) can learn to become calibrated in the training process. They showed that the PLMs in their research had a constant increase in confidence, independent of the accuracy of the predictions. Therefore, it was concluded that PLMs do not learn to be calibrated in training. 

Furthermore, \citet{Valdenegro-Toro_2021_CVPR} presented a meta-analysis of real-world applications that use computer vision. In this research, it is shown that most computer vision applications do not use any form of uncertainty estimation. If they do, it is generally a miscalibrated or only a partial estimation of the uncertainty.

As mentioned, \citet{martino2023exploring} focused on the calibration of ChatGPT, which was based on GPT-3, specifically for sentiment analysis and common sense reasoning. Since the release of GPT-3.5 and GPT-4, along with other LLMs, there is a gap in understanding their uncertainty estimation capabilities. This study aims to build on Pelucchi's work by expanding the evaluation to include multiple LLMs and a broader range of NLP tasks. Furthermore, as shown by \citet{Valdenegro-Toro_2021_CVPR}, uncertainty quantification is often ignored in computer vision applications. Since GPT-4V and Gemini Pro Vision have just been released, little to no research has been done yet on their ability of uncertainty estimation for image recognition tasks. 

Despite existing research, there is a lack of a comprehensive overview of the current state-of-the-art LLMs and VLMs' uncertainty estimation capabilities. This study aims to fill this gap and extend the relatively scarcely researched topic of uncertainty estimation for LLMs and VLMs.

\section{Evaluation Approach}

\subsection{Models and Tasks}
To explore the research questions, this study analyzed four LLMs — GPT-4, GPT-3.5, LLaMA-2-70b, and PaLM 2 — and two VLMs, specifically GPT-4V and Gemini Pro Vision. The selection of these models is aimed at a comprehensive assessment of uncertainty estimation in both LLMs and VLMs. GPT-4 was selected for its leading performance in the LLM domain, serving as a benchmark for comparison. GPT-3.5, LLaMA-2-70b, and PaLM 2 were included due to their notable capabilities and contributions to advancements in the field, offering a diversified perspective of state-of-the-art LLMs. LLaMA-2-70b, being an open-source model, adds value by potentially facilitating further research into enhancing uncertainty estimation in LLMs. The inclusion of GPT-4V and Gemini Pro Vision in the study is particularly significant. These VLMs, being newly released, have not yet been extensively researched, especially in the realm of their uncertainty estimation capabilities. 

LLMs were tested on three distinct NLP tasks to ensure diversity in task complexity and nature: sentiment analysis (SA), math word problems (MP), and named-entity recognition (NER). 

VLMs were tested on one image recognition (IR) task on a new dataset. This dataset is newly created for this study. A more detailed explanation of this dataset will be discussed in Section \ref{sec:jus_dataset}. 

\begin{figure*}[t]
    \centering
    \caption{Synthetic calibration plots demonstrating the interpretation of NCE. All bin sizes are equal. Note how ECE does not indicate direction of miscalibration (overconfidente or underconfident), while NCE does.}
    \label{tab:nce}    
    \begin{subfigure}{0.30\textwidth}
        \includegraphics[width=\linewidth]{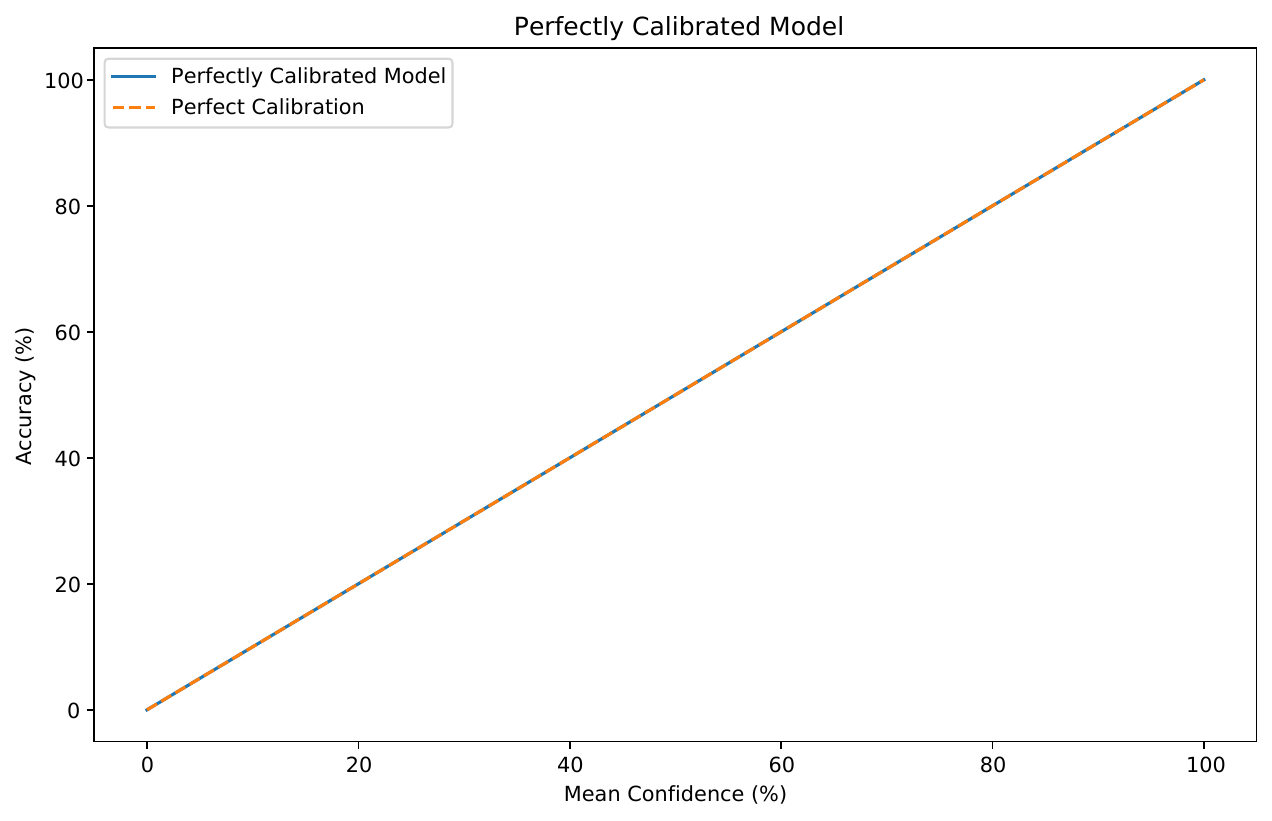}
        \caption{ECE = 0.0, NCE = 0.0}
    \end{subfigure}
    \begin{subfigure}{0.30\textwidth}
        \includegraphics[width=\linewidth]{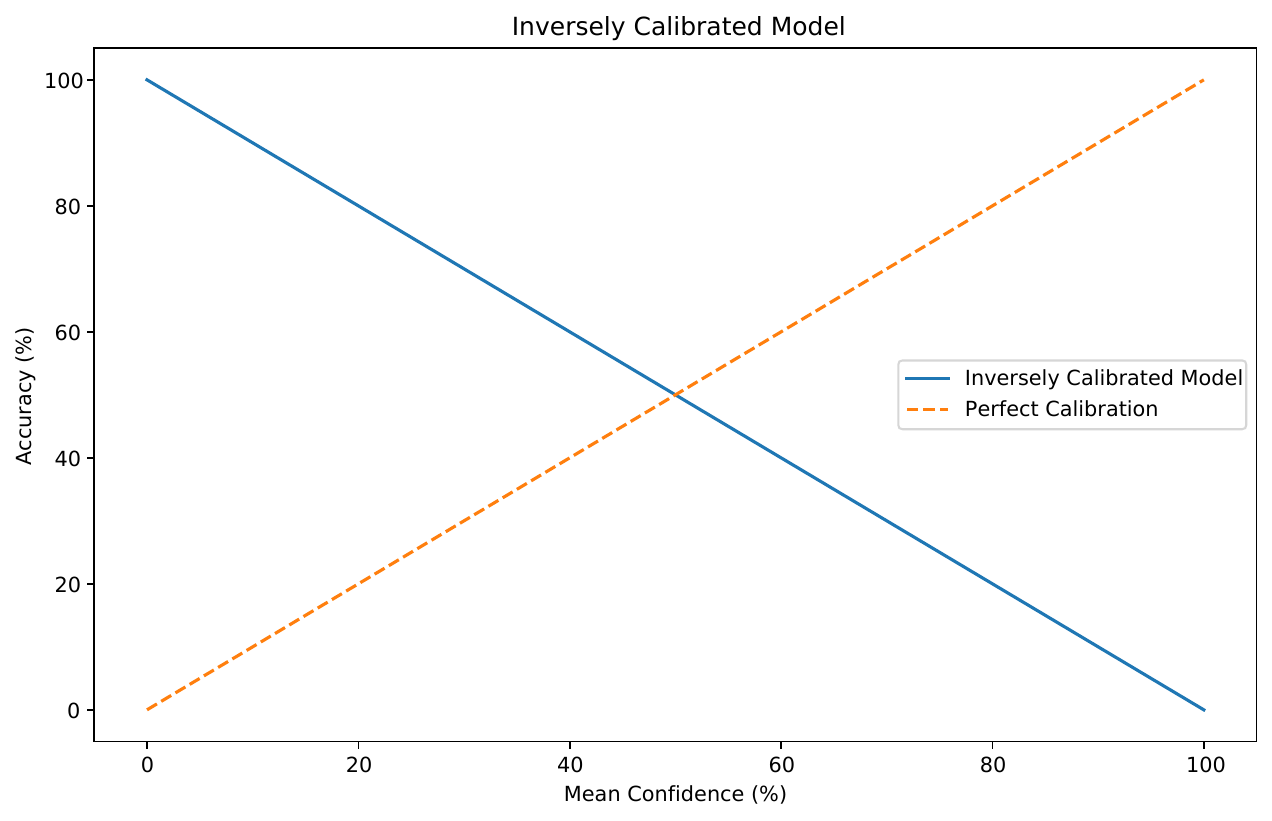}
        \caption{ECE = 60.0, NCE = 0.0}
    \end{subfigure}
    \begin{subfigure}{0.30\textwidth}
        \includegraphics[width=\linewidth]{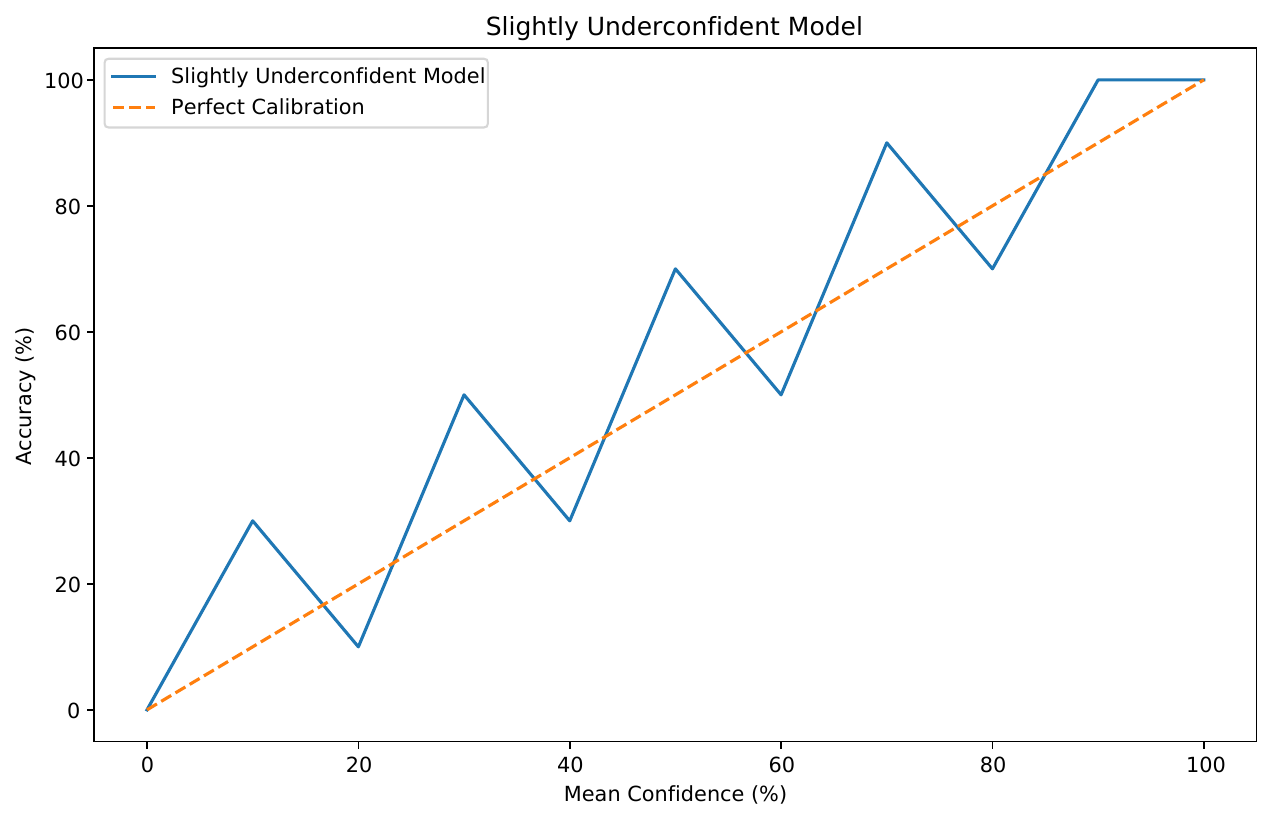}
        \caption{ECE = 13.0, NCE = 5.0}
    \end{subfigure}
    
    \begin{subfigure}{0.30\textwidth}
        \includegraphics[width=\linewidth]{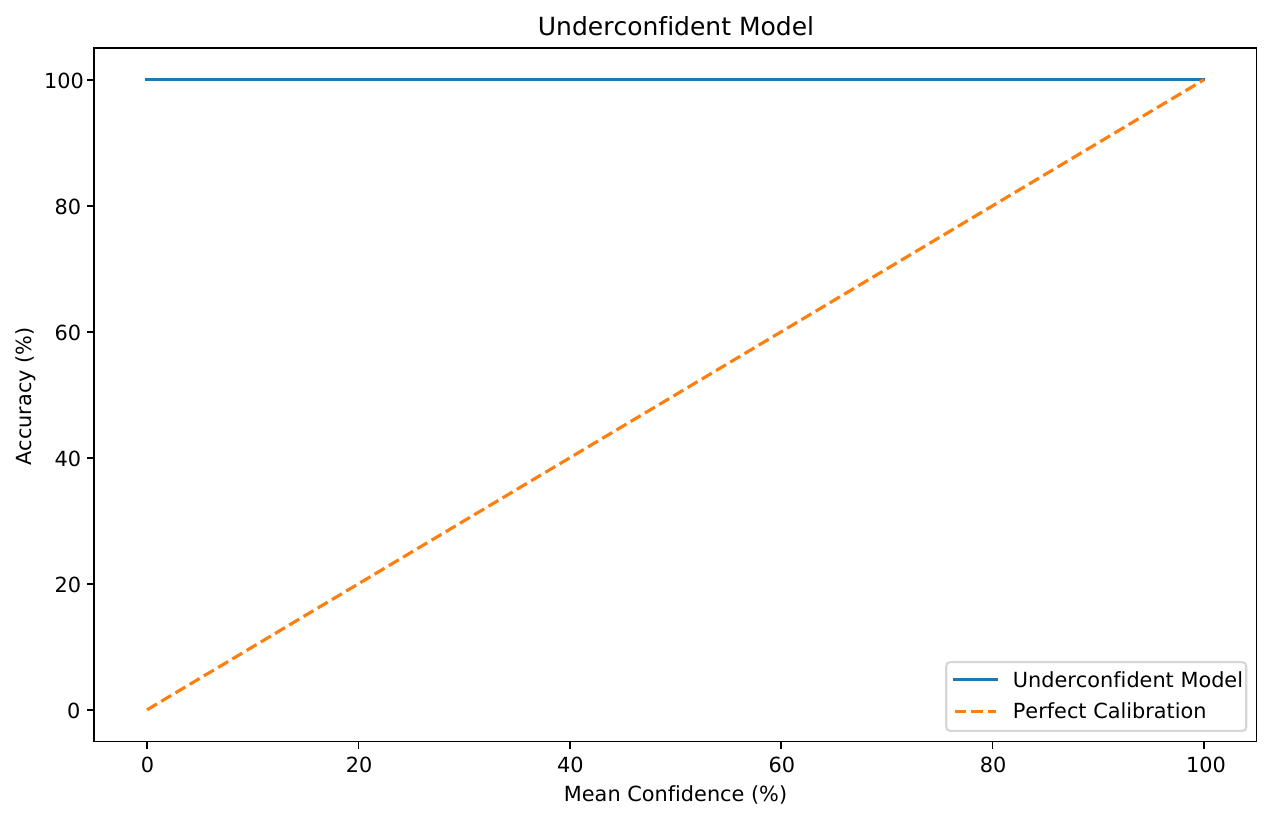}
        \caption{ECE = 55.0, NCE = 55.0}
    \end{subfigure}
    \begin{subfigure}{0.30\textwidth}
        \includegraphics[width=\linewidth]{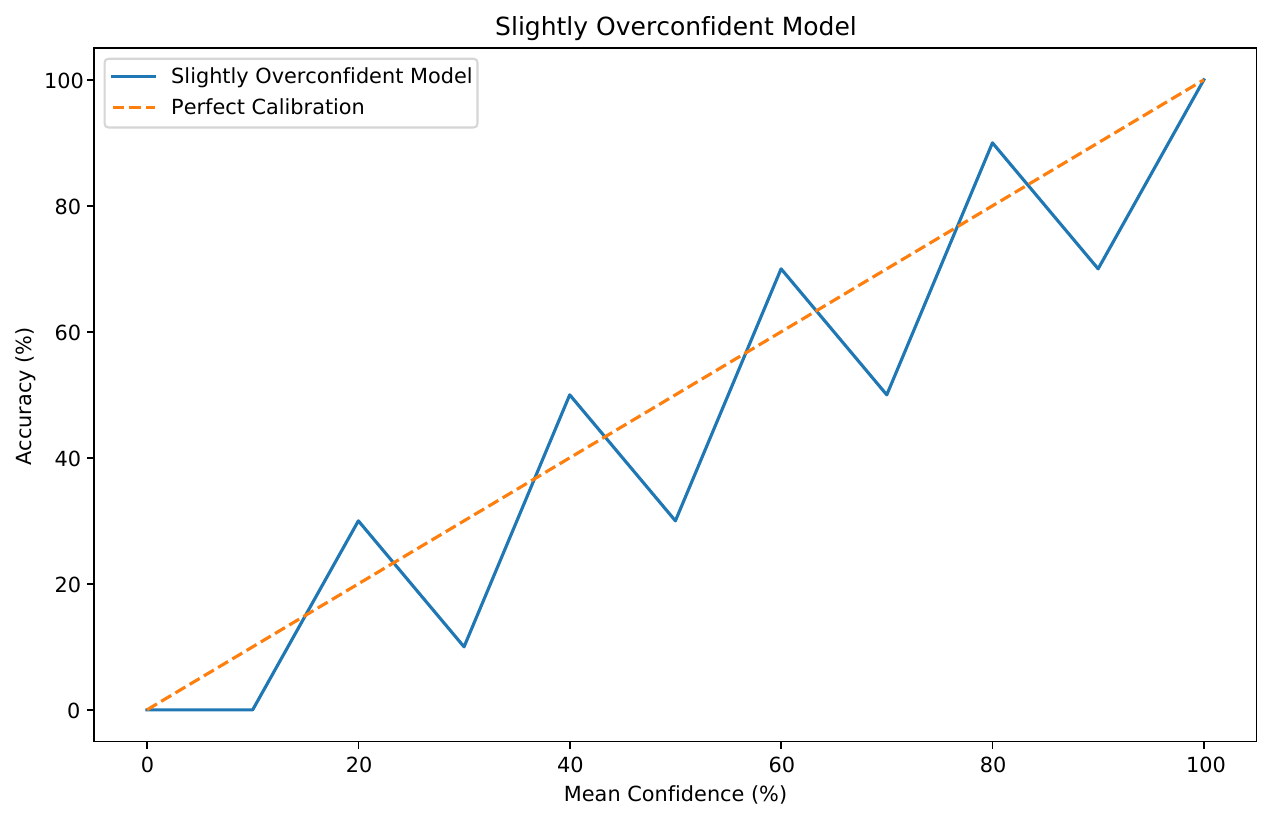}
        \caption{ECE = 13.0, NCE = -5.0}
    \end{subfigure}
    \begin{subfigure}{0.30\textwidth}
        \includegraphics[width=\linewidth]{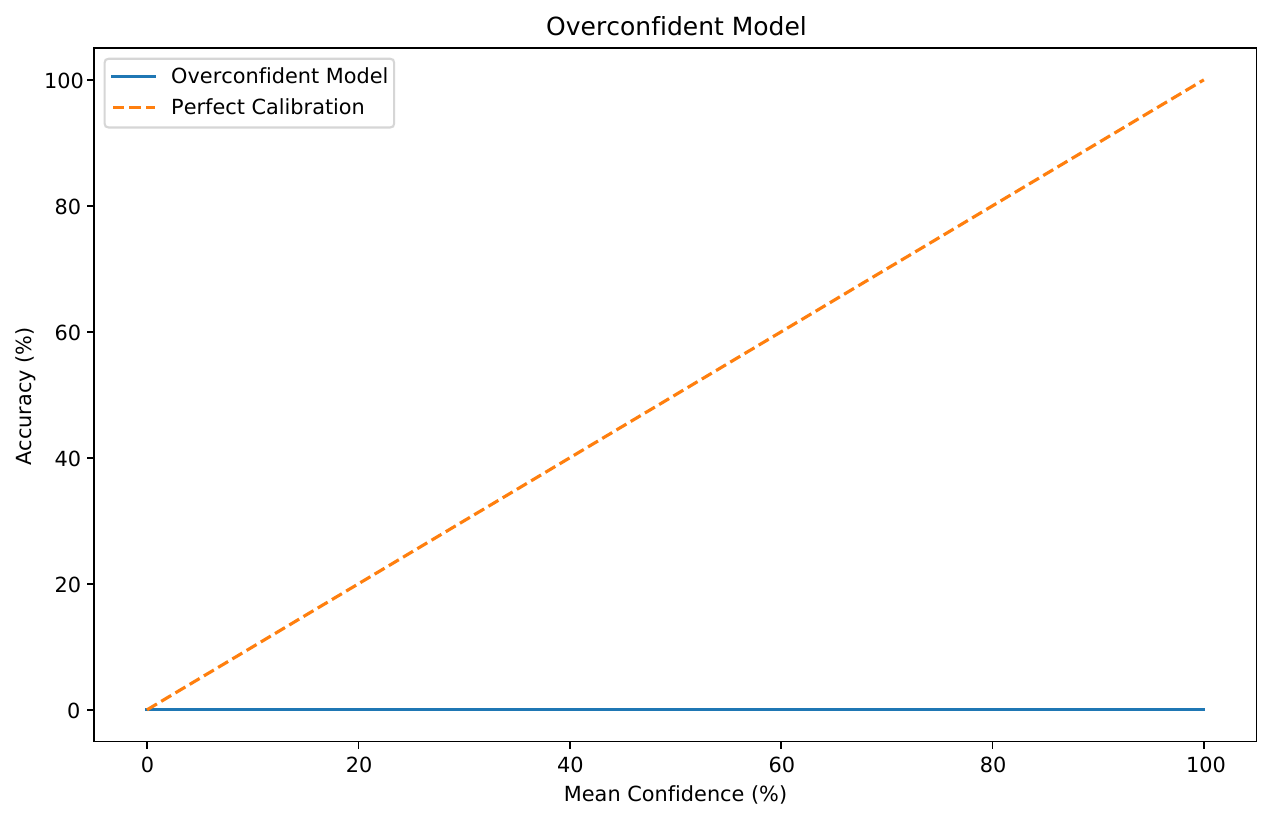}
        \caption{ECE = 55.0, NCE = -55.0}
    \end{subfigure}
\end{figure*}

\subsection{Datasets}
For each task, a corresponding dataset was selected. Each dataset was found in Papers With Code and downloaded from Hugging Face.

For sentiment analysis, the Stanford Sentiment Treebank (SST) dataset \citep{socher-etal-2013-recursive} was used. This research utilizes both the SST2 dataset with binary labels (positive or negative) and the original SST dataset, where sentences are labeled with float values indicating their positivity. The use of these two datasets enables an exploration of various methods of uncertainty estimation. 

Furthermore the GSM8K dataset \citep{cobbe2021gsm8k} was used for the math word problems task and the CoNLL 2003 dataset \citep{tjong-kim-sang-de-meulder-2003-introduction} was used for the named-entity recognition task. The CoNNL 2003 dataset consists of sentences in two languages, English and German. For this research, we focused exclusively on English sentences. From each dataset, 100 random samples were selected for analysis.

\subsubsection{Japanese Uncertain Scenes Image Dataset}
\label{sec:jus_dataset}
Finally, a new dataset was created for the image recognition task, called Japanese Uncertain Scenes (JUS). This dataset consists of 39 images with corresponding prompts. The prompts contain questions about the images, where the questions range from tasks like counting the number of objects or people in an image to identifying the geographical location depicted. All photos were taken in Japan (Osaka, Tokyo, Kyoto, Hiroshima specifically). This dataset was directly created to challenge and test the capabilities of uncertainty estimation in VLMs, with difficult to answer prompts which should be reflected in (increased) verbalized uncertainty. Images were sourced privately, so the exact images are not part of VLM training sets. The full dataset can be seen in Section \ref{sup:dataset} of the Appendix.

The dataset is publicly available at \url{https://github.com/ML-RUG/jus-dataset}.

\subsection{Data Gathering}
The details of all instruction prompts utilized in this study are available in Section \ref{sup:prompt_eng} of the Appendix.

The data was gathered by first prompting the instructions to the models and then prompting the questions. Batch sizes varied based on the task. For sentiment analysis, the models analyzed up to five sentences per batch, speeding up the process of data gathering. However, the models could only process one question at a time for the other tasks. The instruction prompts were reiterated every 10 iterations to maintain consistency in model responses. This repetition was necessary as the models tended to overlook specific instructions if not periodically reminded. All experiments were conducted in December of 2023.

Both LLaMA-2-70b and PaLM 2 could not perform the named-entity task appropriately, requiring multiple instruction prompts per question. Therefore, it was decided to exclude these two models from this task to have a fair comparison, as other LLMs performed well with a single instruction.

Furthermore, for the image recognition task, a new chat was made in GPT-4V for every prompt. This was done to prevent the model from using information from previous prompts.  For instance, if a prior prompt involved an image taken in Japan, the model might use this context to identify subsequent images. In contrast, Gemini Pro Vision did not have memory capabilities at the time of this study. Therefore, creating a separate chat for each prompt for this model was not required.

\begin{table*}[t]
    \centering
    \begin{tabular}{lllllllllllll}
        \toprule
        & \multicolumn{4}{c}{\textbf{Binary Sentiment Analysis}} & \multicolumn{4}{c}{\textbf{Math Word Probs}} & \multicolumn{4}{c}{\textbf{Named Entity Recognition}}\\
        \textbf{Model} & \textbf{Acc} & \textbf{Conf} &\textbf{ECE} & \textbf{NCE} & \textbf{Acc} & \textbf{Conf} &\textbf{ECE} & \textbf{NCE} & \textbf{Acc} & \textbf{Conf} &\textbf{ECE} & \textbf{NCE}\\
        & (\%) & (\%) & & & (\%) & (\%) & & & (\%) & (\%) \\
        \midrule
        
        GPT-4 & 92.0 & 78.5 & 13.5 &  13.5 & 93.0 & 99.8 & 7.20 & -6.80 & 95.3 & 97.9 & 2.53 & -2.58\\ 
        GPT-3.5 & 77.0 & 76.9 & 3.55 & 0.150 & 25.0 & 99.8 & 74.8 & -74.8 & 82.7 & 95.5 & 12.7 & -12.7\\ 
        LLaMA2 & 91.0 & 80.6 & 13.4 & 10.4 & 43.0 & 94.7 & 51.7 & -51.7 & NA & NA & NA & NA\\ 
        PaLM 2 & 90.0 & 79.4 & 14.0 & 10.6 & 56.0 & 99.6 & 43.6 & -43.6 & NA & NA & NA & NA\\
        \bottomrule
    \end{tabular}
    \caption{Summary table for the NLP tasks, presenting mean accuracy, mean confidence, ECE, and NCE. GPT-4 overall demonstrates the smallest ECE and NCE values, suggesting superior calibration relative to other models.
    LLaMA2 corresponds to the 70B variant.}
    \label{tab:sa_mp_ner}
\end{table*}

\begin{table}
    \centering
    \begin{tabular}{lccc}
        \toprule
        \textbf{Model} & \textbf{MAE} & \textbf{MSE} & \textbf{R-Squared} \\ 
        \midrule
        GPT-4 & 0.086 & 0.012 & 0.83 \\ 
        GPT-3.5 & 0.094 & 0.015 & 0.79 \\ 
        LLaMA-2-70b & 0.14 & 0.031 & 0.55 \\ 
        PaLM 2 & 0.12 & 0.027 & 0.61 \\
        \bottomrule
    \end{tabular}
    \caption{Summary table for the float sentiment analysis task, presenting the mean absolute error (MAE), mean squared error (MSE), and the R-squared value.}
    \label{tab:sa_float}
\end{table}

\subsection{Calibration Errors}
\label{sec:calib_errors}

To assess the performance of LLMs, a calibration plot and a confidence density histogram are used. Typically the Expected Calibration Error (ECE) is used, but this metric does not directly reflect over/underconfidence, and we would like to evaluate the direction of miscalibration in each task, as it can be different depending on model and task. For this purpose we introduce the Net Calibration Error (NCE), which can be positive or negative, assessing underconfidence and overconfidence correspondingly. This is shown in Figure \ref{tab:nce}.

In the calibration plots, the error bars are calculated using the normal approximation interval or Wald interval \citep{wallis2013binomial}. This approach was selected due to the binomial nature of the experimental data. A characteristic of the normal approximation interval is to narrow the interval to zero width when the accuracy approaches 0\% or 100\%. Additionally, the width of the interval becomes zero in cases where a confidence bin contains only a single data point. For the calibration plots, answers were grouped in ten confidence bins. This bin size was selected to maintain a balance between having a sufficient number of data points in most bins and ensuring the graph's smoothness. 

The bins of the confidence density histograms were also split up into correct and incorrect answers. By computing the density of these answers in each bin, a deeper understanding of the model's calibration can be obtained. 

Finally, alongside the established ECE and Maximum Calibration Error (MCE), we introduce the Net Calibration Error (NCE) as a novel metric in our analysis. These metrics, including the mean accuracy and mean confidence, were computed for each model across different tasks. 

The ECE is a metric that can be used to assess calibration quality, as it takes the weighted average of the absolute difference between the accuracy and confidence \citep{guo2017calibration}. The ECE is calculated with Eq \ref{eq:ece}:
\begin{equation}
    \label{eq:ece} 
    \text{ECE} = M^{-1} \sum_{m=1}^{M} |B_m| \left| \text{acc}(B_m) - \text{conf}(B_m) \right|
\end{equation}
Where $M$ is the number of bins, \(|B_m|\) is the number of samples whose confidences fall into bin \(m\), \(N\) is the total number of samples, $\text{acc}(B_m)$ is the accuracy (between 0-100\%) of the predictions in bin \(m\), and $\text{conf}(B_m)$ is the mean confidence (between 0-100\%) of the predictions in bin \(m\).

The MCE and NCE are two variations of the ECE. The MCE shows the absolute maximum difference between the predicted confidence and actual accuracy for any of the bins and is calculated with equation \ref{eq:mce} \citep{guo2017calibration}:
\begin{equation}\label{eq:mce} 
    \text{MCE} = \max_{m} \left| \text{acc}(B_m) - \text{conf}(B_m) \right|
\end{equation}
In this paper, we introduce the NCE. The NCE closely resembles the ECE. The only difference is that the NCE uses the weighted average of the straightforward difference between the accuracy and the confidence, rather than their absolute difference, as can be seen in equation \ref{eq:nce}: 

\begin{equation}\label{eq:nce} 
    \text{NCE} = M^{-1} \sum_{m=1}^{M} |B_m| \left(\text{acc}(B_m) - \text{conf}(B_m)\right) 
\end{equation}

This approach allows the NCE to indicate the direction of miscalibration, a feature not offered by either the ECE or the MCE. 
Despite its novelty and current lack of adoption in scientific literature, we argue that the NCE provides essential insights absent in the ECE and MCE. However, it is important to note that the NCE alone does not reflect calibration quality, as an NCE of zero can occur even with poor calibration. This limitation is mitigated by the ECE, which already quantifies the degree of miscalibration. Therefore, the ECE, MCE, and NCE collectively provide a comprehensive overview of model calibration, showing the magnitude, direction, and maximum of the miscalibration. In Section \ref{supp:interp_nce} of the Appendix, we provide further demonstration of the interpretation of the NCE.

\section{Experimental Results}
\label{sec:results}

\subsection{Large Language Models}

\begin{figure}[t]
    \centering
    \begin{subfigure}{0.49\linewidth}
        \includegraphics[width=\textwidth]{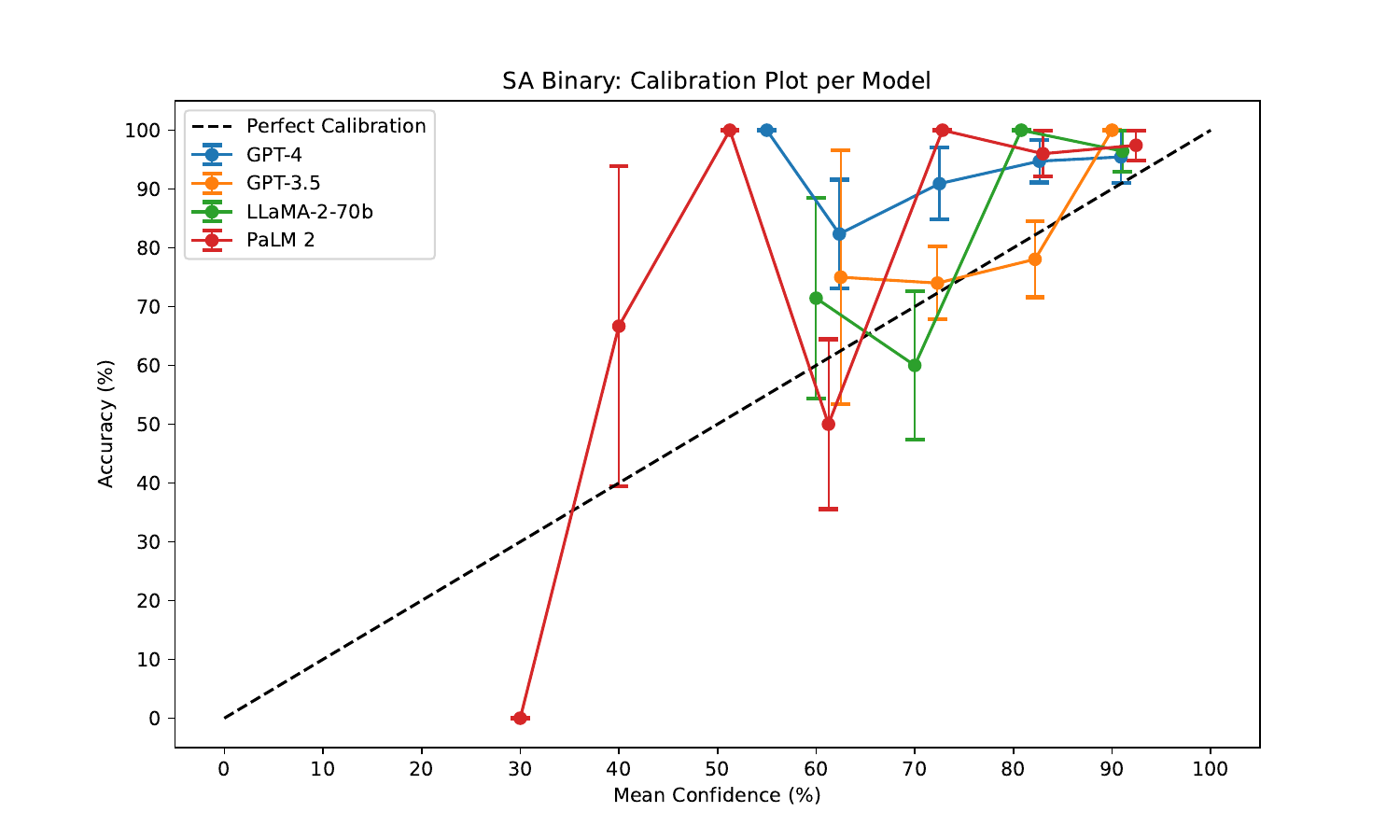}
    \end{subfigure}
    \begin{subfigure}{0.49\linewidth}
        \includegraphics[width=\textwidth]{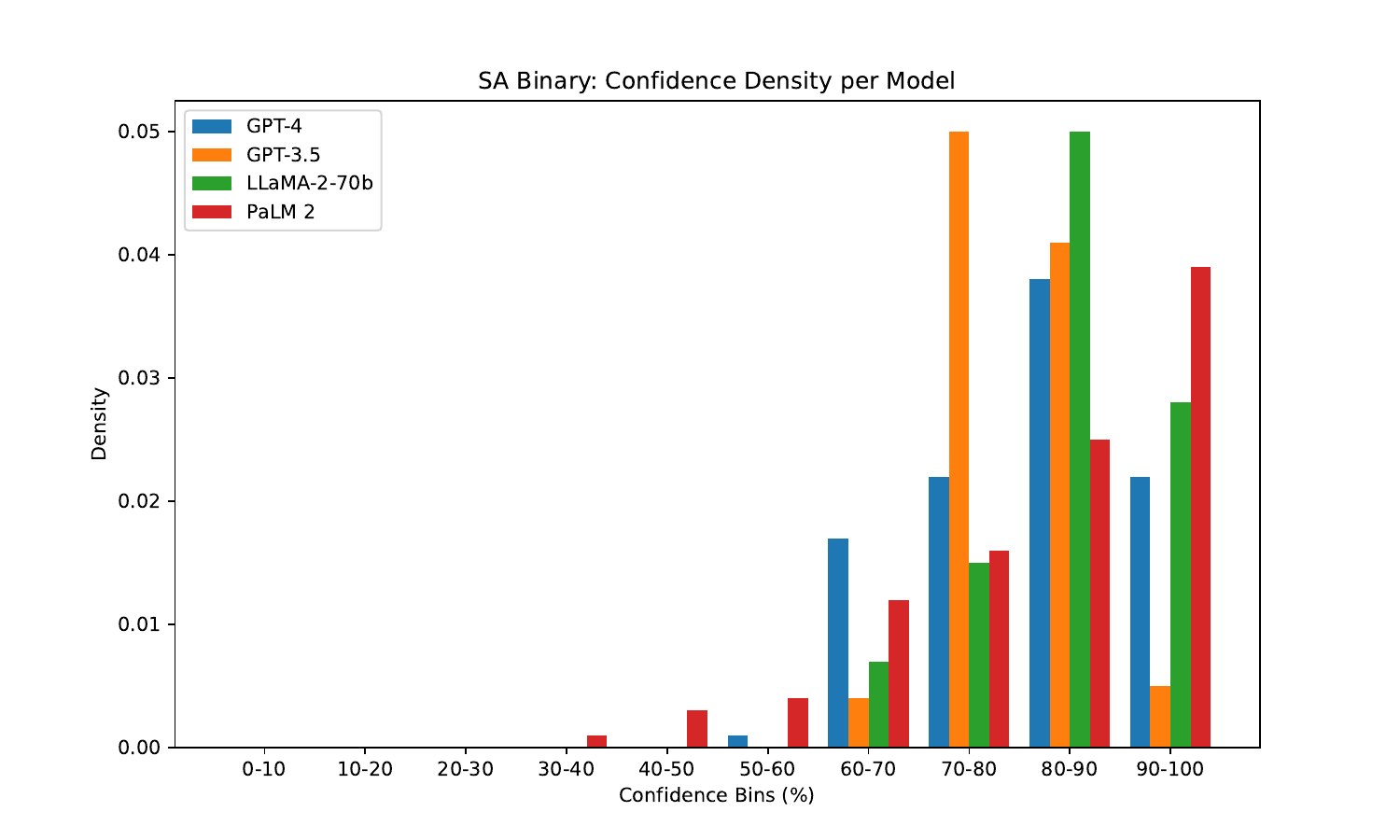}
    \end{subfigure}    
    \caption{Calibration plots and confidence histograms for the sentiment analysis task with binary labels. GPT-3.5 shows closer calibration to the ideal, whereas the other models mostly exhibit underconfidence.}
    \label{fig:sa_calibration}
\end{figure}

\subsubsection{Sentiment Analysis}

Figure \ref{fig:sa_calibration} shows the calibration plot for the sentiment analysis task with binary labels. GPT-3.5 exhibits the closest alignment to the diagonal line. The diagonal line represents perfect calibration, where the confidences match the accuracies. In contrast, the other models generally demonstrate higher accuracy than their reported confidence, signifying a tendency toward underconfidence.

This underconfidence is further illustrated in Table \ref{tab:sa_mp_ner}. The Table shows that despite GPT-4's high correctness rate, it often reports lower confidence levels. In contrast, GPT-3.5 shows better calibration where its mean accuracy and mean confidence differ by only 0.1\%. Nonetheless, the ECE suggests minor miscalibration, with the average deviation being 3.55\%, which is notably lower compared to the other models. Furthermore, it can be seen that the NCE is positive for all models, confirming the underconfidence.

Additionally, Table \ref{tab:sa_float} shows the results of the model performances on the sentiment analysis task with float labels. GPT-4 emerges as the most accurate model, with the lowest MAE at 0.086 and MSE at 0.012. Its R-squared value of 0.83 signifies a high level of predictive accuracy, indicating that GPT-4's predictions closely align with the actual outcomes. GPT-3.5 follows closely, demonstrating good uncertainty estimation capabilities, although slightly less precise than GPT-4. LLaMA-2-70b and PaLM 2, while competent, show greater errors and lower R-squared values, suggesting room for improvement in their calibration processes.

\subsubsection{Math Word Problems}

\begin{figure}[t]
    \centering
    \begin{subfigure}{0.49\linewidth}
        \includegraphics[width=\textwidth]{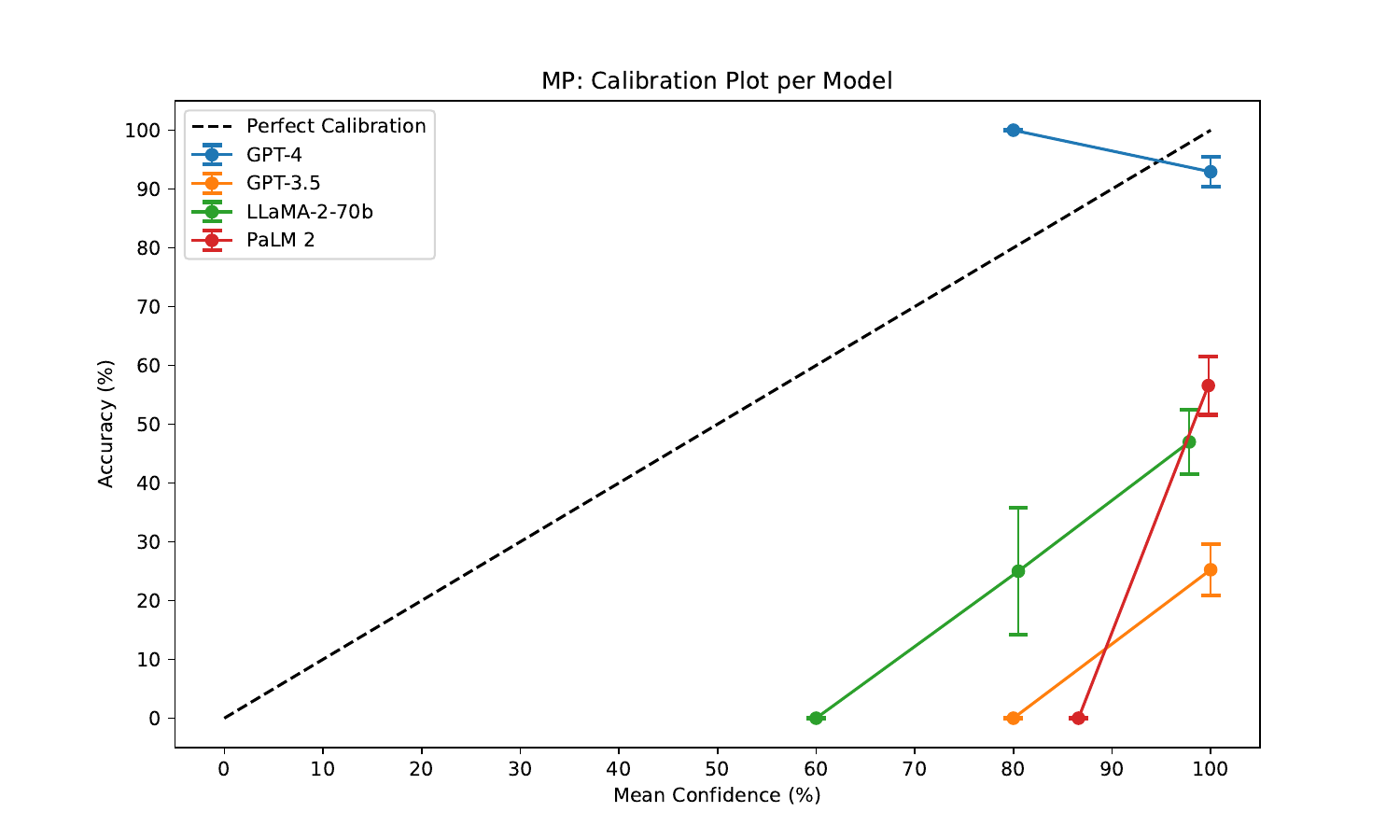}
    \end{subfigure}
    \begin{subfigure}{0.49\linewidth}
        \includegraphics[width=\textwidth]{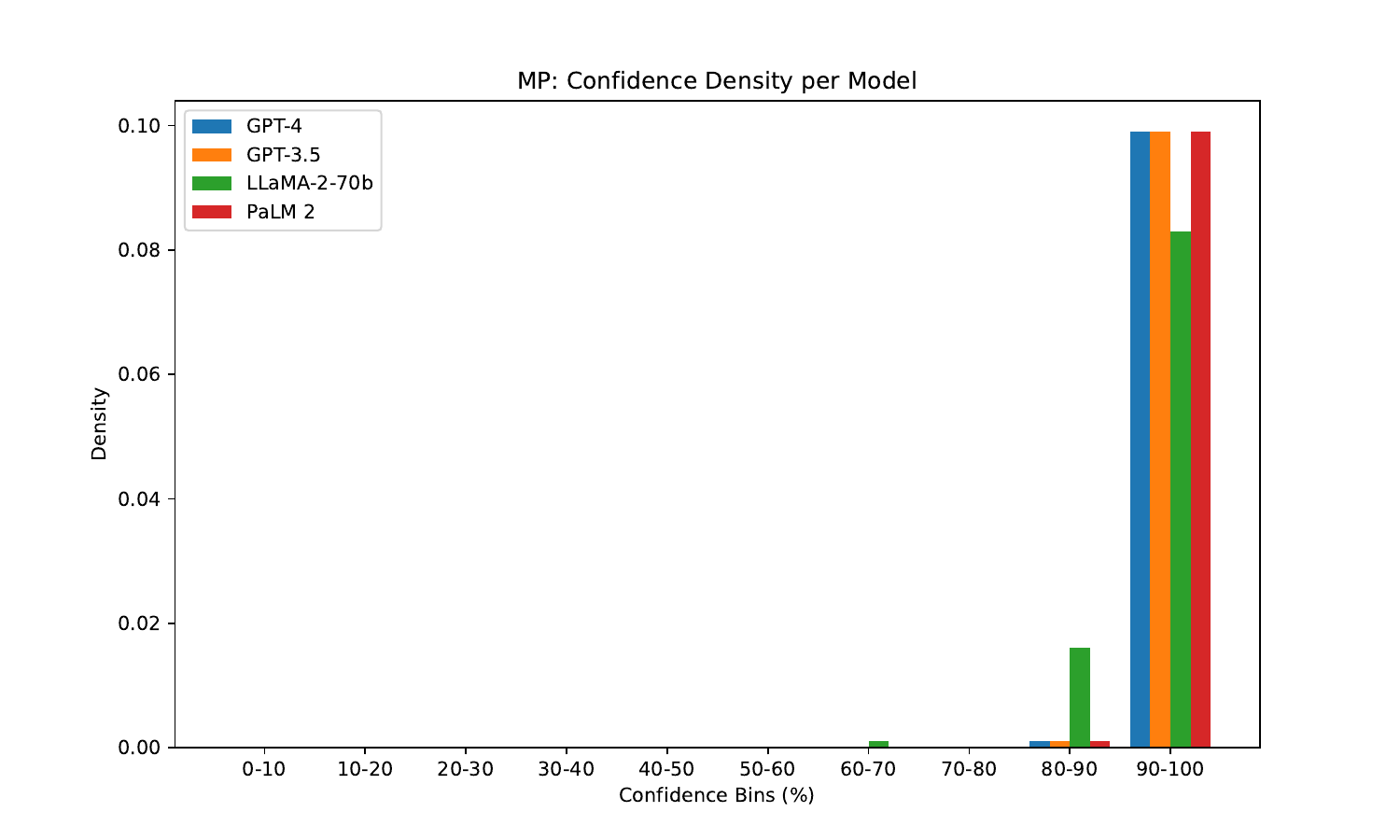}
    \end{subfigure}            
    \caption{Calibration plots and confidence histograms for the math word problems task. All models exhibit excessive overconfidence except for GPT-4, and all models output extremely high confidence in their answers.}
    \label{fig:mp_calibration}
\end{figure}

Figure \ref{fig:mp_calibration} displays the calibration plot for the math word problems task. Except for GPT-4, all models exhibit excessive overconfidence, as shown by their positioning well below the diagonal line. GPT-4 stands out as the only model that appears to be well-calibrated for this task. Figure \ref{fig:mp_calibration} further demonstrates that all models show extremely high confidence, with almost all outputted confidences falling in the 90-100\% confidence bin. Table \ref{tab:sa_mp_ner} shows that only GPT-4 can justify this high confidence, whereas all the other models cannot. This is particularly true for GPT-3.5, which has an ECE of 74.8\% and a corresponding NCE of -74.8\%, indicating that all confidence bins show underconfidence, where the average deviation from the diagonal line is 74.8\%. Moreover, PaLM 2 exhibits the highest MCE at 86.6.

\subsubsection{Named-Entity Recognition}

\begin{figure}[t]
    \centering
    \begin{subfigure}{0.49\linewidth}
        \includegraphics[width=\textwidth]{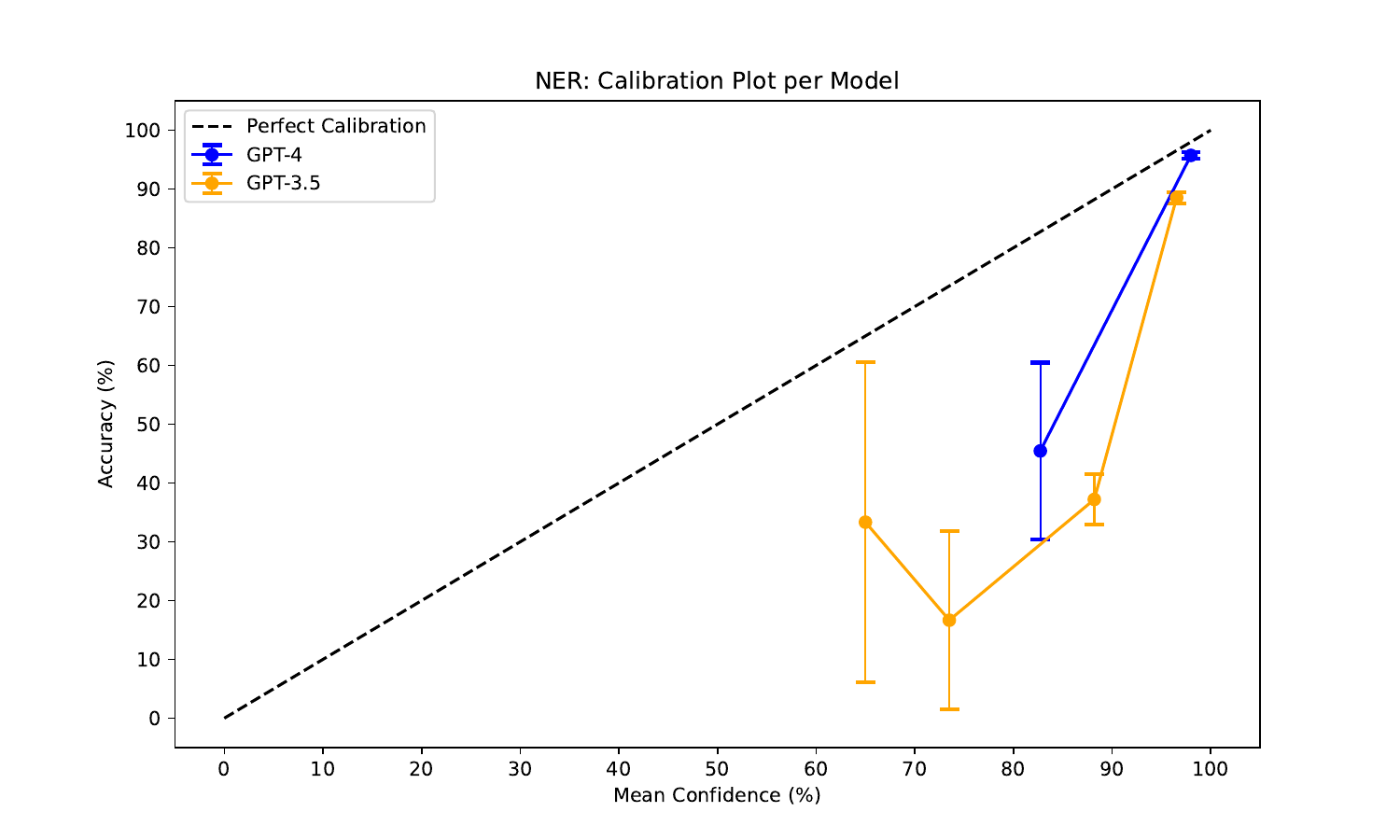}
    \end{subfigure}
    \begin{subfigure}{0.49\linewidth}
        \includegraphics[width=\textwidth]{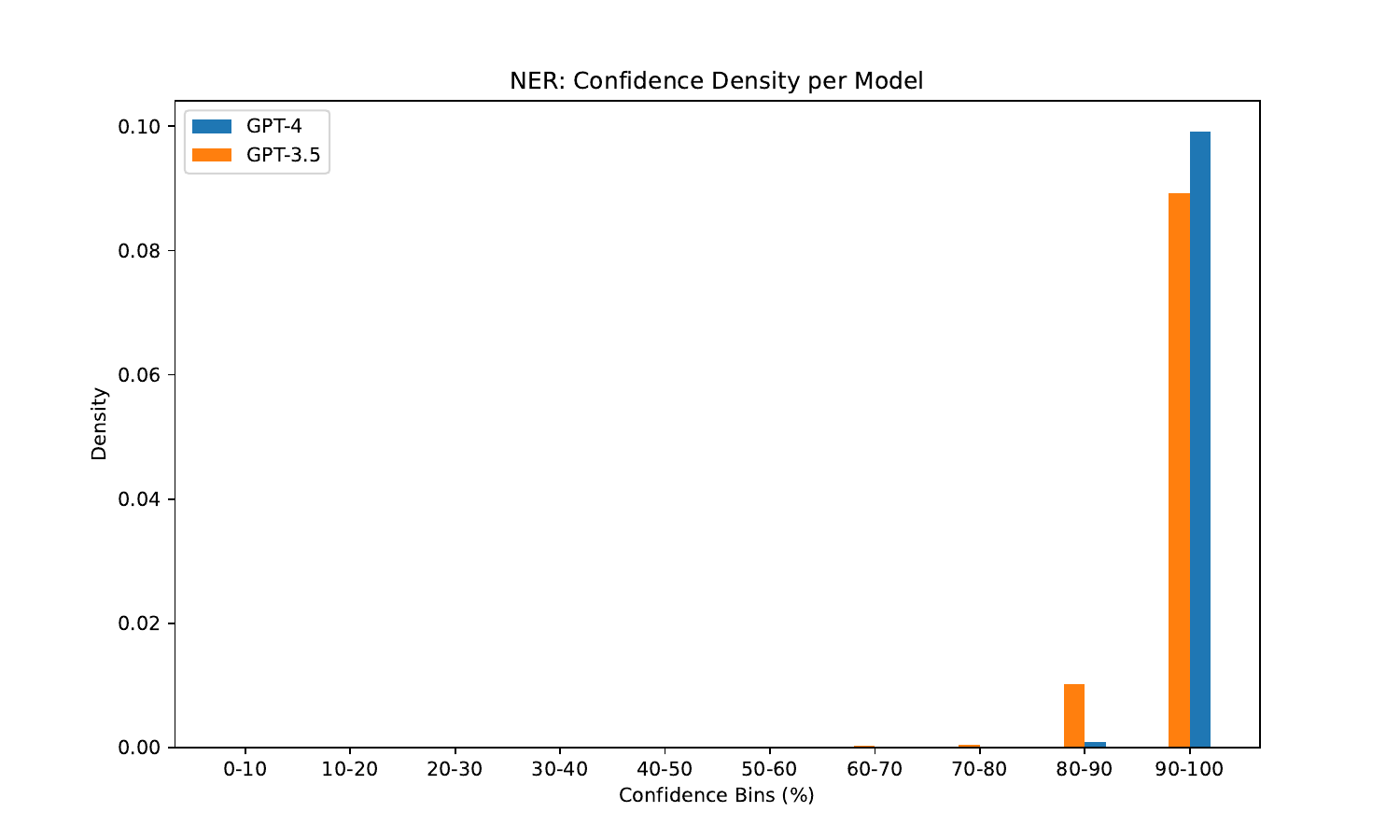}
    \end{subfigure}        
    \caption{Calibration plots and confidence histograms for the named-entity recognition task. GPT-4 seems to be better calibrated than GPT-3.5, although both models show overconfidence.}
    \label{fig:ner_calibration}
\end{figure}

The calibration plot for the named-entity recognition task is shown in Figure \ref{fig:ner_calibration}. As mentioned in the Methods section, PaLM 2 and LLaMA-2-70b were not capable of performing this task and therefore only GPT-4 and GPT-3.5 were evaluated. Despite both models showing overconfidence again, GPT-3.5 seems to be more overconfident compared to its successor. Interestingly, Figure \ref{fig:ner_calibration} reveals that GPT-4 actually exhibited higher confidence levels than GPT-3.5. However, due to GPT-4's superior accuracy, its overconfidence is lower. This distinction is further supported by the data in Table \ref{tab:sa_mp_ner} where both models exhibit a negative NCE, indicative of overconfidence. Notably, GPT-4 is, on average, approximately 10\% less overconfident than GPT-3.5.

\subsection{Vision Language Models}
To evaluate the VLMs, a calibration plot together with confidence density histograms was made. Additionally, also the ECE, MCE, NCE, mean confidence and mean accuracy were calculated. 

Alternative instruction prompts for evaluating VLMs were also created for this study. For the instruction prompts, analysis, and example answers of this method, please refer to Sections \ref{sup:prompt_eng} and \ref{sup:example_answers} in the Appendix.

\subsubsection{Image Recognition on JUS}
\begin{figure}[t]
    \centering
    \begin{subfigure}{0.49\linewidth}
        \includegraphics[width=\textwidth]{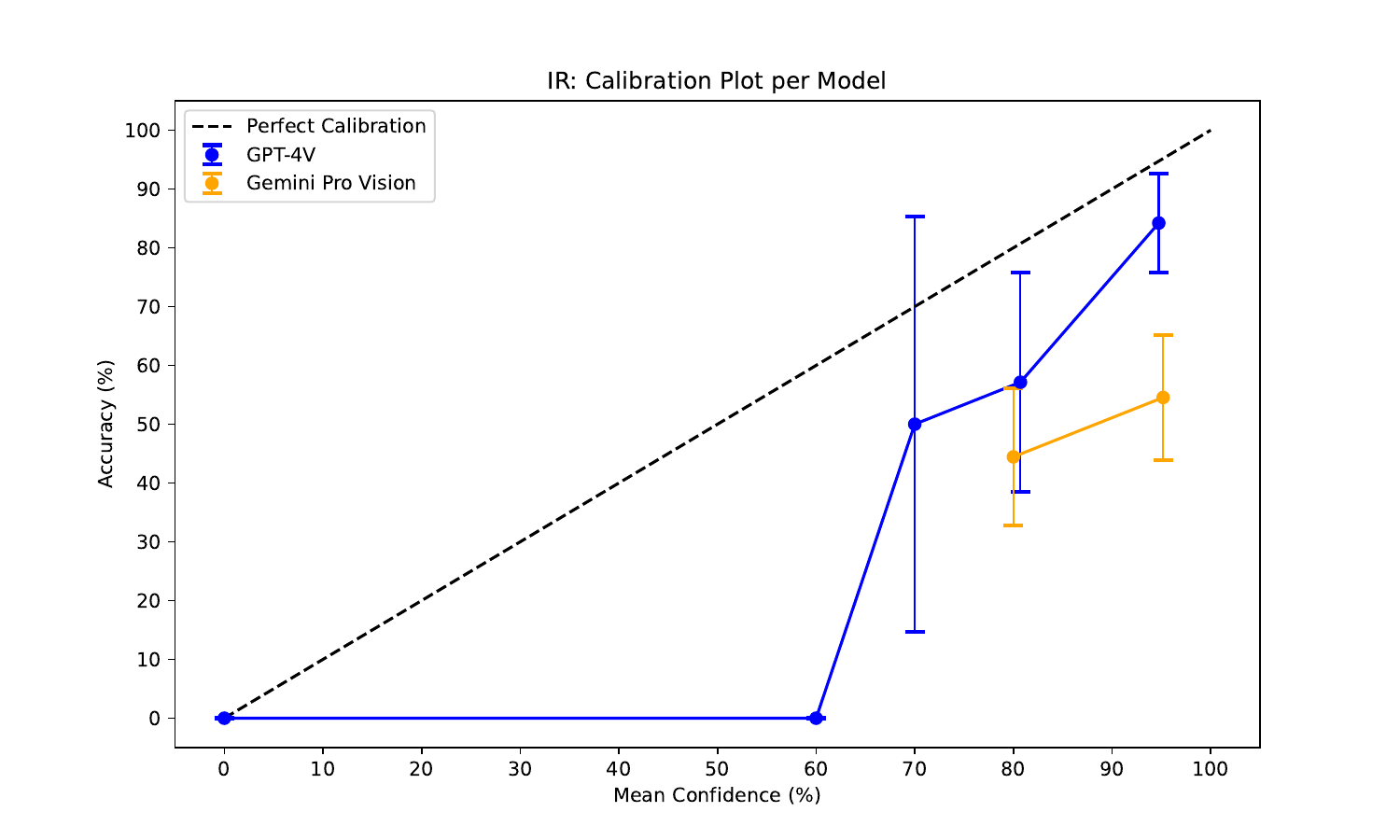}
    \end{subfigure}
    \begin{subfigure}{0.49\linewidth}
        \includegraphics[width=\textwidth]{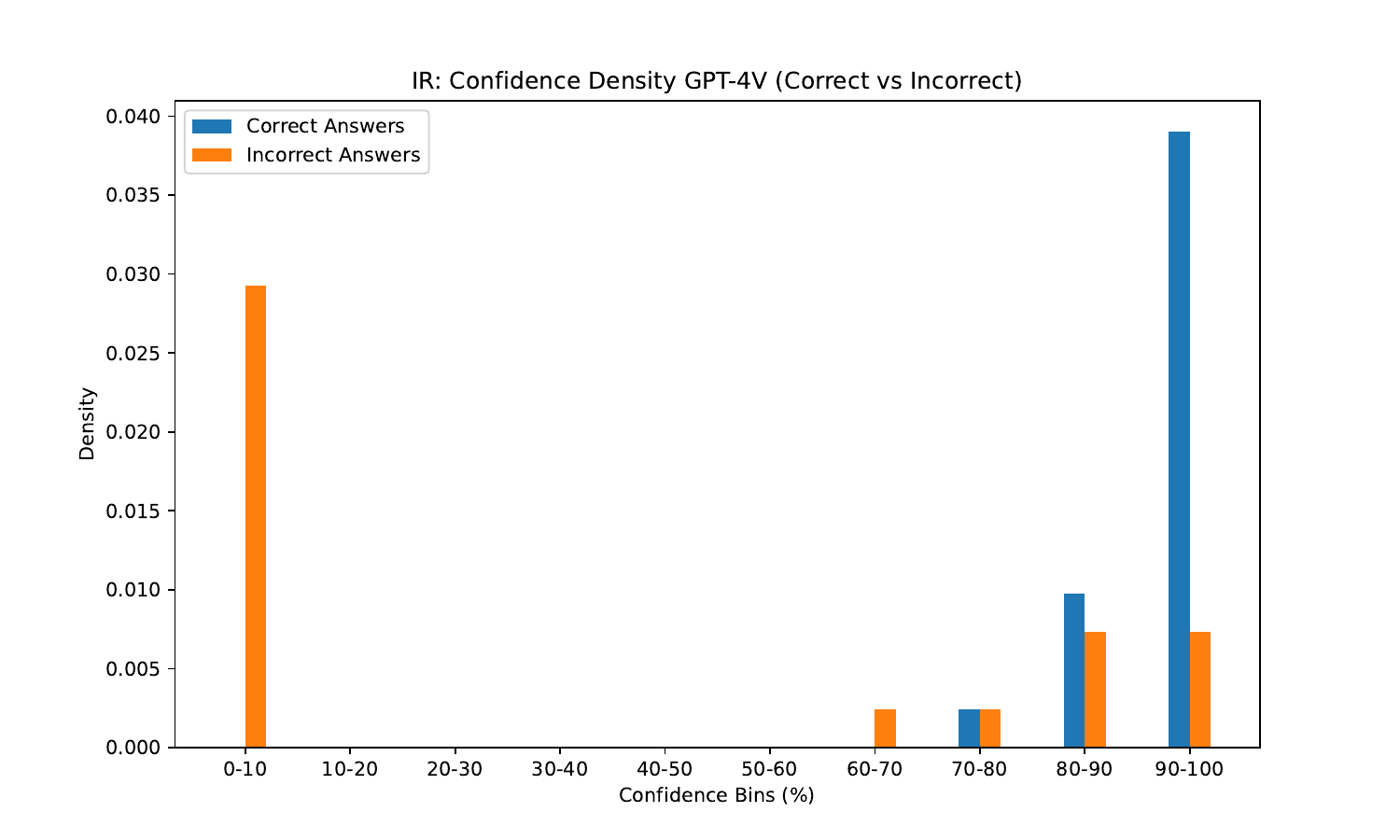}
    \end{subfigure}
    \caption{Calibration plot and confidence density histogram for VLM image recognition on JUS. GPT-4V shows superior performance over Gemini Pro Vision.}
    \label{fig:vlm_calibration}
\end{figure}

In Figure \ref{fig:vlm_calibration}, the calibration plot for the image recognition task reveals that GPT-4V is more closely aligned with the diagonal line, indicating superior performance over Gemini Pro Vision, although both models exhibit overconfidence. Notably, GPT-4V achieves perfect calibration in instances where both its mean confidence and actual accuracy are zero. 

An example of GPT-4's 0\% confidence output is presented in Figure \ref{tab:vlm_dataset_results}. This answer prompt demonstrates that the model is aware of its inability to provide the correct answer, and therefore outputs 0\% confidence and does not give an answer to the question, showing perfect calibration. In contrast, Gemini Pro Vision provides an incorrect answer with a confidence level of 80\%, showing very poor calibration. Additional example answers are provided in Section \ref{sup:example_answers} of the Appendix.

This discrepancy in calibration quality is further demonstrated in Table \ref{tab:vlm}. GPT-4 has an ECE of 11.3, which is markedly lower than Gemini Pro Vision's ECE of 38.4. The negative NCE values for both models underscore their tendency towards overconfidence. 

Tables \ref{tab:vlm_mean_std} and \ref{tab:vlm_conf_interval} in the Appendix present results for six images with a counting prompt (regression), and both mean/std and 95\% confidence interval uncertainties do not faithfully represent model uncertainty, being almost random.

\begin{table}[t]
\centering
\begin{tabular}{llllll}
\toprule
\textbf{Model} & \textbf{Acc} & \textbf{Conf} & \textbf{ECE} & \textbf{NCE} \\
 & (\%) & (\%) \\
\midrule
GPT-4 & 51.2 & 62.6 & 11.3 & -11.3 \\
Gemini & 50.0 & 88.4 & 38.4 & -38.4 \\
Pro Vision \\
\bottomrule
\end{tabular}
\caption{Summary for VLM image recognition on JUS, presenting mean accuracy, mean confidence, ECE, MCE, and NCE. GPT-4V shows superior calibration compared to Gemini Pro Vision, while both are overconfident.}
\label{tab:vlm}
\end{table}

\section{Discussion}
\label{sec:discussion}

A primary observation is the generally poor accuracy of LLMs in estimating their own uncertainty across different NLP tasks. This inaccuracy is mostly caused by overconfidence, except for the sentiment analysis task where a tendency towards underconfidence was noted. For the math word problems and named-entity recognition tasks, the models displayed alarmingly high confidence levels, with the majority of predictions falling within the 90-100\% confidence interval. This overconfidence is particularly concerning given that, with the exception of GPT-4, the models' actual accuracies did not substantiate such high confidence levels.

GPT-4 demonstrated superior calibration relative to the other LLMs. However, it is worth noting that the model consistently outputted high confidence levels, which, due to its corresponding high accuracy, resulted in a more calibrated performance. This raises the consideration if GPT-4 is genuinely better calibrated, or if this is merely a byproduct of its higher accuracy. 

The VLMs also showed limited accuracy in uncertainty estimation, with a predominant trend toward overconfidence. GPT-4V showed better calibration compared to Gemini Pro Vision. Interestingly, GPT-4V showed a good level of self-awareness, particularly in recognizing instances where it lacked the capabilities to answer a complex question. This self-awareness underscores a significant advancement in VLMs, emphasizing the importance of models recognizing their own limitations as a key component of effective uncertainty estimation.
    
The outcomes of this study align with the conclusions drawn by \cite{martino2023exploring} and \cite{jiang2021can}, which similarly identified a tendency towards overconfidence in LLMs. For this study, a wide range of LLMs have been tested on a variety of NLP tasks, thereby validating the results of previous research across a wider spectrum. Additionally, this study assesses the uncertainty estimation capabilities of recently introduced VLMs. 

\subsection{Limitations}
This study, while providing valuable insights into the uncertainty estimation capabilities of LLMs and VLMs, is subject to several limitations that require consideration. Firstly, to create the calibration plots, data was categorized based on confidence levels. As highlighted in the Results section, the models tended to produce exceedingly high confidence levels despite simultaneously achieving low accuracy scores. This led to an uneven distribution of data across the confidence bins, with some bins having sparse data, thereby introducing variability in the calibration plots. Addressing this challenge requires a greater number of task iterations to ensure all confidence bins have enough data points. However, given the models' tendency to yield high confidence levels for certain tasks, achieving enough data points in all confidence bins could be notably time-consuming. 

Each task was performed once per model. This approach does not account for potential performance variability across different chats. To enhance the reliability of the findings, it would be beneficial to conduct multiple iterations of each task for every model, although this might significantly increase the time and resources required for the study.

We focused on a select group of LLMs and VLMs. While these models are selected to create a comprehensive overview of the current technology, they do not account for the entire landscape of language and vision language models. Tasks requiring more nuanced understanding or complex reasoning may yield different results in terms of uncertainty estimation.

The JUS dataset has a limited size, only 39 images, but we believe it shows fundamental issues with VLM uncertainty estimation and limits of these models, as they seem to be unable to count objects, and performing counting as a regression task, they produce nonsensical and highly miscalibrated confidence intervals.

\section{Conclusions}
\label{sec:conclusions}
In this study we focused on how accurately LLMs estimate their uncertainty accross various NLP tasks. The findings indicate that LLMs generally exhibit poor accuracy in estimating their own uncertainty when performing various natural language processing tasks, with a predominant trend towards overconfidence in their outputs. However, among the LLMs, there is variation in the quality of uncertainty estimation, with GPT-4 exhibiting the highest quality and being the best calibrated.

Interestingly, the type of task influences this estimation accuracy; for instance, in sentiment analysis, models tended to be underconfident, whereas in math word problems and named-entity recognition tasks, a significant overconfidence was observed.

The second research question examined the uncertainty estimation capabilities of VLMs in an image recognition task. Similar to LLMs, the results showed that VLMs demonstrate limited accuracy in self-estimating uncertainty in an image recognition task, trending towards overconfidence. Notably, GPT-4V showed a relatively better calibration when compared to Gemini Pro Vision.

These results provide a foundational basis for future studies. It is shown that the current LLMs and VLMs show poor uncertainty estimation quality. Therefore, it is of high importance to study how uncertainty estimation can be improved. 

\cite{NEURIPS2022_9d560961} showed how 'Chain of Thought' (CoT) prompting can significantly increase the accuracy of LLMs on certain tasks. It would therefore be interesting to see if this CoT-prompting could also improve the uncertainty estimation quality in LLMs and VLMs. 

LLaMA-2-70b is an open-source model. This presents the opportunity for future research to investigate how direct modifications to the model could improve its uncertainty estimation capabilities.

\clearpage
\bibliography{custom}

\clearpage
\appendix
\onecolumn

\section{Broader Impact Statement}
\label{sup:broader_impact}

Uncertainty estimation is an important problem to detect incorrect predictions or plain hallucinations, particularly for large language and vision-language models.

Our paper adds to the growing literature that LLMs and VLMs do not have native capabilities to estimate their own uncertainty, and this could have societal consequences. Users of these models should be aware that their answers/replies can be incorrect, misleading, and verbalized uncertainty will not be a signal to detect this kind of mispredictions, even as that is the common expectations associated to confidences.

More resources should be devoted to build LLMs and VLMs with proper uncertainty estimation, instead of just increasing predictive performance and building larger and larger models. Uncertainty estimation and confidence assessment is part of reasoning and intelligence.

\section{Instruction Prompt Engineering and Detailed Prompts}
\label{sup:prompt_eng}

Besides performing the tasks, the models were also required to express their confidence in each answer. To ensure the models accurately understood both the task and confidence estimation, an instruction prompt was developed for each task. This development involved a trial-and-error process, where each version of the instruction prompts was tested across all models for all tasks. This eventually resulted in the instruction prompt shown below.

The following instruction prompts were all inspired by the study of \cite{martino2023exploring}.

\textbf{Sentiment Analysis Binary}: Hello, I would like you to perform sentiment analysis on 100 English sentences. I will present the sentences per batch, each containing 5 sentences. Please classify each sentence as “Positive” or “Negative”  according to its sentiment. Moreover, please rate your confidence in your answer between 0 and 100\%. The answer should be a list with the format “Sentiment (confidence\%)”. Say “understood” if you have understood. The list of sentences will follow.

\textbf{Sentiment Analysis Float}. Hello, I would like you to perform sentiment analysis on 100 English sentences. Please express its level of positive sentiment as a float value from 0.0 to 1.0. The answer should be in the format “(float value between 0.0 and 1.0)”. Do not add any additional information.  I will split the sentences into 10 batches, consisting of 10 sentences per batch. Say “understood” if you have understood. The batches with sentences will follow.

\textbf{Math Word Problems}. Hello, I would like you to solve math word problems on 100 problems. Please rate your confidence in the answer you give between 0 and 100\%. The answer should be of the format “Answer (confidence\%)”. This should be your only output, so do not provide any intermediate steps that lead to the answer. Say “understood” if you have understood. The math word problems will follow.
     
\textbf{Named-Entity Recognition}. Hello, I would like you to perform named-entity recognition on 100 English sentences. I will provide the sentences one by one. Please classify each word of the sentence with the tags 0-8:
    \begin{itemize}
    \item 'O' (0): This tag represents "Outside" or "Other." It is used for words that do not belong to any named entity.
    \item 'B-PER' (1): "Beginning-Person." This means the word corresponds to the beginning of a person entity.
    \item 'I-PER' (2): "Inside-Person." This means the word is inside a person entity.
    \item 'B-ORG' (3): "Beginning-Organization." This means the word corresponds to the beginning of an organization entity.
    \item 'I-ORG' (4): "Inside-Organization." This means the word is inside an organization entity.
    \item 'B-LOC' (5): "Beginning-Location." This means the word corresponds to the beginning of a location entity.
    \item 'I-LOC' (6): "Inside-Location." This means the word is inside a location entity.
    \item 'B-MISC' (7): "Beginning-Miscellaneous." This means the word is the beginning of a word that does not fall into any of the previous entities (person, organization, location) but does belong to a named entity.
    \item 'I-MISC' (8): "Inside-Miscellaneous." This tag is for words within a miscellaneous entity that are not the beginning word.
    \end{itemize} 
    Moreover, please rate your confidence in the answer you gave between 0 and 100\%. The answer should be a list with the format “[Tag1 (confidence\%), Tag2 (confidence\%), Tag3 (confidence\%), …, Tagn (confidence\%)]” where n is the number of items in the sentence. Say “understood” if you have understood. The list of sentences will follow.
    
\textbf{Image Recognition with Confidence Levels}. \textit{Question prompt...}Moreover, please rate your confidence in your answer between 0 and 100\%. The answer should be in the format: "Answer (confidence\%)".

\textbf{Image Recognition with Mean and Standard Deviation}. \textit{Question prompt...}Please give your actual prediction. Moreover, please express your answer by giving a mean and a standard deviation to reflect the uncertainty in your answer. The answer should be in the format: "Mean = [mean value], SD = [standard deviation value]".

\textbf{Image Recognition with 95\% Confidence Interval}. \textit{Question prompt...}Please give your actual prediction. Moreover, please express your estimate as a 95\% confidence interval. This means you should provide a range within which you are 95\% confident the true value lies. Format your answer as: '[Lower Bound, Upper Bound]', where the lower bound is the start of the range and the upper bound is the end of the range. Ensure that this interval reflects a 95\% confidence level based on your estimation.

\section{Data Samples NLP Tasks}
\label{sup:nlp_task_data}

From each dataset, 100 samples were randomly chosen. This approach allows for a balanced representation of the data, minimizing any potential biases and ensuring that the findings are robust and reliable. The indices listed below, presented in the format [index1, index2, ..., indexn], correspond to the specific samples selected from their respective datasets.

\begin{itemize}
    \item \textbf{Sentiment Analysis Float (SST dataset)}: [1836, 4201, 2287, 2234, 239, 3604, 8243, 1701, 7442, 1792, 1687, 3759, 6429, 4333, 2941, 7422, 3946, 8062, 4199, 1487, 7024, 2129, 963, 2497, 8263, 7466, 3993, 3573, 3987, 1383, 867, 6960, 4554, 6001, 5950, 3360, 7023, 533, 7031, 4806, 4151, 612, 3753, 1107, 4346, 2722, 609, 4887, 7435, 2146, 2009, 625, 3667, 4154, 4328, 5132, 6342, 3097, 4179, 2664, 778, 8048, 4872, 7804, 2612, 940, 5616, 5844, 5244, 2599, 6935, 4344, 1289, 7013, 997, 4952, 8321, 5018, 5533, 3586, 7770, 3250, 721, 7941, 4357, 2147, 186, 2937, 4599, 7971, 5497, 346, 6964, 4786, 7964, 0, 7650, 6765, 6637, 5941]
    \item \textbf{Sentiment Analysis Binary (SST2 dataset)}: [66682, 53090, 56562, 25791, 40181, 29117, 36719, 38196, 25905, 42393, 15702, 50111, 6376, 45138, 36415, 30148, 17086, 56186, 22341, 38297, 47013, 6680, 40122, 8214, 3380, 67284, 16394, 25127, 66964, 20789, 35066, 15417, 2942, 11594, 17135, 13422, 65901, 23825, 63598, 10236, 47065, 51326, 42231, 29513, 48335, 47735, 53725, 32420, 25671, 9305, 21168, 67152, 38343, 20707, 39861, 37870, 61651, 66778, 6520, 29546, 21267, 27350, 46338, 30838, 13950, 15050, 36899, 1990, 49030, 31455, 7910, 17991, 52228, 32968, 20973, 11075, 53731, 28329, 12122, 21189, 48020, 25860, 64088, 36555, 65124, 8146, 11319, 14651, 47224, 48922, 37303, 54210, 33568, 30623, 36127, 35318, 10640, 60563, 38968, 35300]
    \item \textbf{Math Word Problems (GSM8K dataset)}: [5913, 5926, 726, 2227, 2405, 570, 3155, 6656, 7457, 2303, 7323, 5236, 526, 751, 2150, 1415, 1782, 2563, 7288, 5970, 770, 4170, 1879, 3063, 2917, 4027, 1818, 4926, 1848, 657, 29, 3796, 5497, 2338, 1013, 6783, 4605, 977, 4851, 1236, 337, 6597, 3866, 248, 1735, 70, 3820, 4641, 4905, 5604, 1010, 4612, 3631, 867, 2659, 27, 281, 6707, 7339, 6207, 4184, 319, 7084, 5702, 3406, 6215, 3207, 3245, 3563, 656, 6104, 1447, 7370, 5782, 806, 4981, 5814, 3066, 6035, 6158, 6686, 574, 5564, 4738, 1816, 6239, 6259, 1405, 1765, 6918, 627, 1499, 5699, 6398, 913, 4343, 601, 304, 4559, 3203]
    \item \textbf{Named-Entity Recognition (CoNLL 2003 dataset)}: [7535, 10543, 10718, 678, 7396, 8147, 3010, 8671, 3382, 6381, 167, 304, 565, 9616, 9326, 1478, 5240, 14004, 9739, 9987, 4261, 2383, 6648, 3054, 7476, 3407, 13646, 2262, 3387, 2046, 9521, 781, 6502, 260, 10637, 5171, 1123, 13843, 7538, 2691, 3737, 1310, 1180, 8034, 8496, 4168, 10161, 6065, 1290, 7393, 5260, 12075, 8112, 79, 10710, 7278, 1769, 3757, 5863, 12450, 12366, 6341, 3624, 6438, 12542, 4822, 13379, 7138, 11467, 4503, 5540, 8394, 12438, 3914, 1707, 8321, 12402, 7738, 6396, 11977, 11815, 7464, 3025, 13477, 3455, 10899, 11416, 5905, 11266, 2161, 13066, 7842, 10067, 11767, 1898, 8306, 5703, 820, 7739, 1543]
\end{itemize}

\section{Interpretation Net Calibration Error}
\label{supp:interp_nce}

Table \ref{tab:nce} presents six synthetic plots to demonstrate the interpretation of the NCE. The first row features two plots with an NCE of zero, implying neither overconfidence nor underconfidence. However, it does not say anything about the models' calibration levels. The ECE clarifies this: 0 for the left plot, signifying perfect calibration, and 60 for the right plot, indicating significant miscalibration. The right plot maintains an NCE of zero because the levels of underconfidence and overconfidence are balanced, effectively neutralizing each other and yielding an NCE of zero. Consequently, an NCE of zero is interpreted as indicating no trend towards either overconfidence or underconfidence. 

The second row depicts plots with a positive NCE. A positive NCE indicates that, on average, the accuracy is higher than the confidence, and therefore the model tends towards underconfidence. The NCE shows that the model is slightly underconfident, with an average of 5\% above the perfect calibration line. The ECE indicates an average miscalibration of 13\%. 

The right plot shows a model that has 100\% accuracy across all confidence bins. Interestingly, the ECE and NCE are equal. This indicates complete underconfidence, with all data points on or above the diagonal line, meaning that the accuracy is consistently equal to or higher than the confidence. In this case, the average miscalibration is 55\%, where all miscalibration is due to underconfidence. 

In the third row, plots with a negative NCE are displayed. A negative NCE indicates that, on average, the accuracy is lower than the confidence, and therefore the model tends towards overconfidence. The left plot mirrors the one above, showing mild overconfidence with an average deviation of 5\% below the ideal calibration line.

The right plot shows a model which has an accuracy of 0\% across all confidence bins. Interestingly, the NCE is the negative counterpart of the ECE. This indicates complete overconfidence, with all data points lying on or below the diagonal line, meaning that the accuracy is consistently equal to or lower than the confidence. In this case, the average miscalibration is 55\%, where all miscalibration is due to overconfidence. 

From these observations, we can deduce the following about the NCE:
\begin{itemize}
    \item \textbf{\(NCE = 0\)}: No trend towards over- or underconfidence.
    \item \textbf{\(NCE > 0\)}: Model tends towards underconfidence.
    \item \textbf{\(NCE < 0\)}: Model tends towards overconfidence.
    \item\(NCE = ECE\) where \(ECE \neq 0\): Complete underconfidence, with all data points at or above the ideal calibration line.
    \item \(-NCE = ECE\) where \(ECE \neq 0\): Complete overconfidence, with all data points at or below the ideal calibration line.
\end{itemize} 

\newpage

\section{Pearson Correlation Tests}
A Pearson Correlation Test was performed to check the correlation between accuracy and  mean confidence per confidence bin. These results are presented in Tables \ref{tab:pct_1}, \ref{tab:pct_2}, \ref{tab:pct_3}, and \ref{tab:pct_4} mostly show high p-values. This is probably caused by the relatively low number of confidence bins that contained any data points.

\begin{table}[h!]
    \centering
    \captionof{table}{Results for the Pearson Correlation Test on the sentiment analysis binary task.}
    \label{tab:pct_1}
    \begin{tabular}{lcc}
        \hline
        \textbf{Model} & \textbf{Correlation Coefficient} & \textbf{p-value} \\ \hline
        GPT-4 & 0.126 & 0.840 \\ 
        GPT-3.5 & 0.801 & 0.199 \\ 
        LLaMA-2-70b & 0.774 & 0.226 \\ 
        PaLM 2 & 0.725 & 0.0654 \\ \hline
    \end{tabular}

    \captionof{table}{Results for the Pearson Correlation Test on the math word problems task.}
    \label{tab:pct_2}
    \begin{tabular}{lcc}
        \hline
        \textbf{Model} & \textbf{Correlation Coefficient} & \textbf{p-value} \\ \hline
        GPT-4 & -1.0 & 1.0 \\ 
        GPT-3.5 & 1.0 & 1.0 \\ 
        LLaMA-2-70b & 1.0 & 0.0072 \\ 
        PaLM 2 & 1.0 & 1.0 \\ \hline
    \end{tabular}
    \captionof{table}{Results for the Pearson Correlation Test on the named-entity recognition task.}
    \label{tab:pct_3}
    \begin{tabular}{lcc}
        \hline
        \textbf{Model} & \textbf{Correlation Coefficient} & \textbf{p-value} \\ \hline
        GPT-4 & 1.0 & 1.0 \\ 
        GPT-3.5 & 0.77 & 0.23 \\ \hline
    \end{tabular}
    \captionof{table}{Results for the Pearson Correlation Test on the image recognition task.}
    \label{tab:pct_4}
    \begin{tabular}{lcc}
        \hline
        \textbf{Model} & \textbf{Correlation Coefficient} & \textbf{p-value} \\ \hline
        GPT-4 & 0.81 & 0.10 \\ 
        Gemini Pro Vision & 1.0 & 1.0 \\ \hline
    \end{tabular}
\end{table}        

\FloatBarrier
\section{Japanese Uncertain Scenes Image Recognition Dataset}
\label{sup:dataset}

In this section, the complete image recognition dataset is presented. The difficulty of the prompts is intentionally designed to evaluate how challenging tasks affect the models' uncertainty estimations. Furthermore, the dataset includes trick questions and other challenging prompts where obtaining the answer is difficult. Ultimately, the purpose of the dataset is not to assess the accuracy of specific models but to compare their calibration levels. 

Each image is paired with its associated prompt and the correct answer. In cases where an image corresponds to two prompts, they are differentiated as (a) for the first prompt and (b) for the second prompt. Please note that these prompts were presented separately to the VLMs. Prompts 2, 3, 9, 10, 16, and 17 were used for the image recognition task with standard deviation and mean, and the 95\% confidence interval as the required output.  

The images in this dataset were obtained from private sources, copyright is owned by Matias Valdenegro-Toro, the images are not available on the Internet\footnote{Previous to public release of this dataset.}. The purpose of using privately owned images is to prevent that VLMs would have these images on their training sets. Photographs were taken in Tokyo, Kyoto, Osaka, Hiroshima, and Fujikawaguchiko.

Images were labelled by the authors, in the context of Tobias Groot's Bachelor Thesis. Labels correspond to prompts and correct answers, and answers were validated by experts on Japan.

\begin{table*}[h!]
    \centering
    \captionof{figure}{Image recognition dataset prompts 1-6}
    \label{tab:dataset1}
    \begin{tabular}{ccc}

        \includegraphics[width=0.3\textwidth]{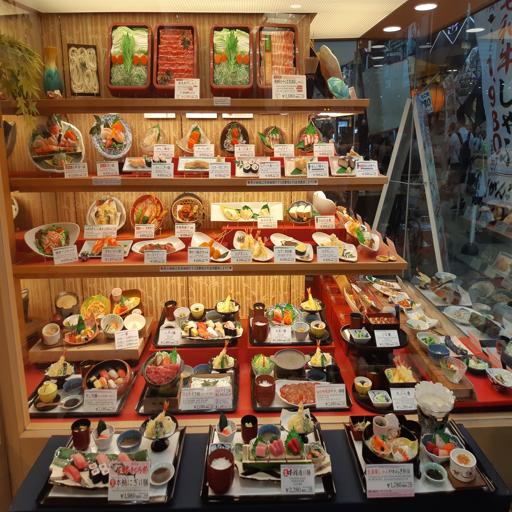} &
        \includegraphics[width=0.3\textwidth]{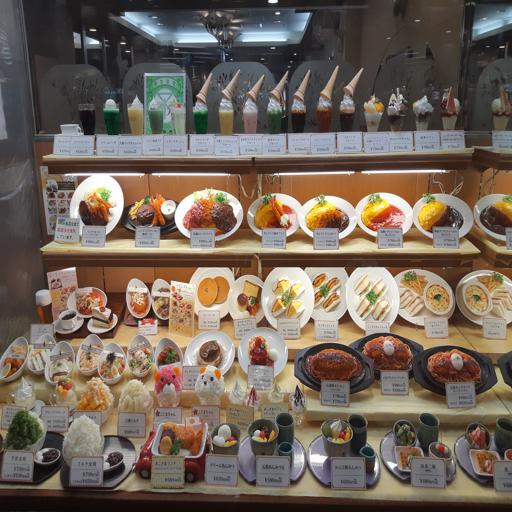} &
        \includegraphics[width=0.3\textwidth]{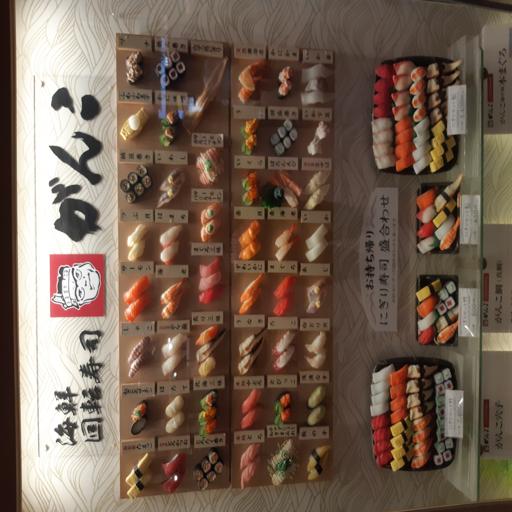} \\
        \parbox[t]{0.3\textwidth}{ \textbf{1. Prompt:} How many food items are shown in this photo? Moreover, please rate your confidence in your answer between 0 and 100\%. The answer should be in the format: "Answer (confidence\%)".} &
        \parbox[t]{0.3\textwidth}{ \textbf{2. Prompt:}  How many desserts are shown in this photo? Moreover, please rate your confidence in your answer between 0 and 100\%. The answer should be in the format: "Answer (confidence\%)".} &
        \parbox[t]{0.3\textwidth}{ \textbf{3. Prompt:}   How many sushi pieces are displayed here? Moreover, please rate your confidence in your answer between 0 and 100\%. The answer should be in the format: "Answer (confidence\%)". } \\

        \multicolumn{3}{c}{} \\

        \parbox[t]{0.3\textwidth}{ \textbf{Correct Answer:} Unknown, there are around 50 meals/plates, but a lot more food items. The ideal answer of the model would be to have 0\% confidence and not give a prediction.} &
        \parbox[t]{0.3\textwidth}{ \textbf{Correct Answer:}  20-30} &
        \parbox[t]{0.3\textwidth}{ \textbf{Correct Answer:}  201} \\

        \multicolumn{3}{c}{} \\

        \includegraphics[width=0.3\textwidth]{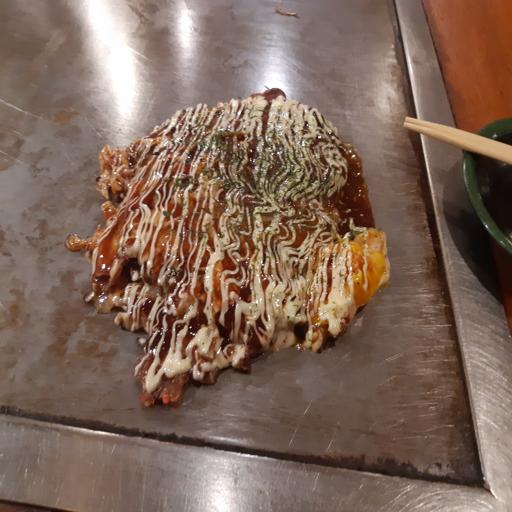} &
        \includegraphics[width=0.3\textwidth, angle=+90]{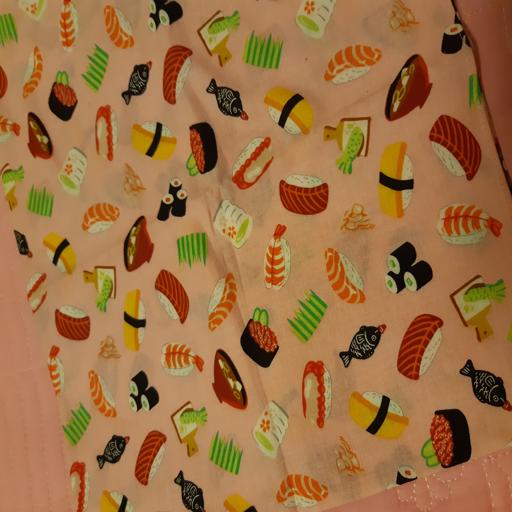} &
        \includegraphics[width=0.3\textwidth]{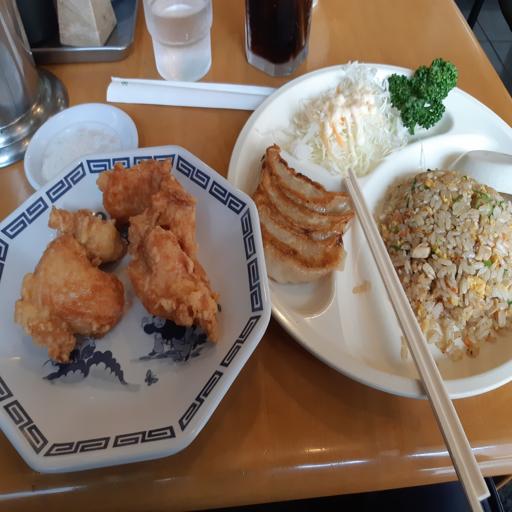} \\
        \parbox[t]{0.3\textwidth}{ \textbf{4. Prompt:} What is shown in this photo? Moreover, please rate your confidence in your answer between 0 and 100\%. The answer should be in the format: "Answer (confidence\%)".} & \parbox[t]{0.3\textwidth}{ \textbf{5. Prompt:} How many sushi pieces are in this photo? Moreover, please rate your confidence in your answer between 0 and 100\%. The answer should be in the format: "Answer (confidence\%)". 
         } & \parbox[t]{0.3\textwidth}{ \textbf{6. Prompt:} What kind of food is presented in this photo? Moreover, please rate your confidence in your answer between 0 and 100\%. The answer should be in the format: "Answer (confidence\%)". } \\

        \multicolumn{3}{c}{} \\

        \parbox[t]{0.3\textwidth}{ \textbf{Correct Answer:} Okonomiyaki} &
        \parbox[t]{0.3\textwidth}{ \textbf{Correct Answer:}  Right answer would notice that these are sushi drawings.} &
        \parbox[t]{0.3\textwidth}{ \textbf{Correct Answer:}  Egg fried rice, fried chicken, and gyoza. Japanese food is also correct.} \\        
    \end{tabular}
\end{table*}

\begin{table*}[h!]
    \centering
    \captionof{figure}{Image recognition dataset prompts 7-12}
    \label{tab:dataset2}
    \begin{tabular}{ccc}

        \includegraphics[width=0.3\textwidth]{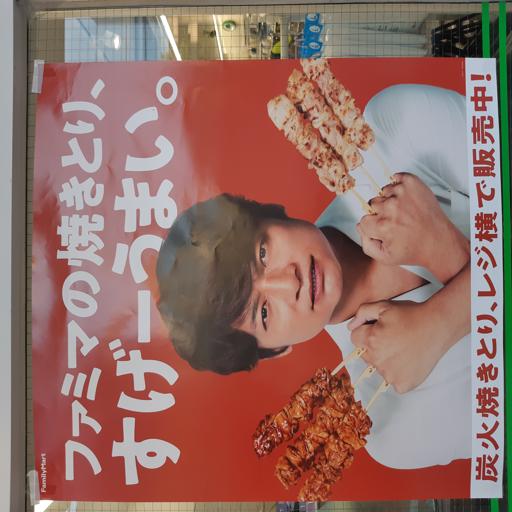} &
        \includegraphics[width=0.3\textwidth, angle=+90]{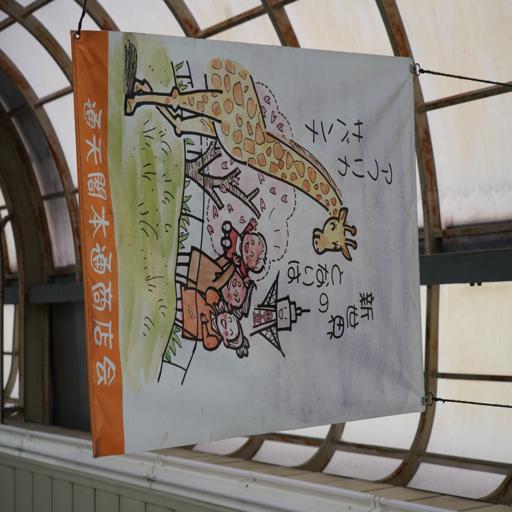} &
        \includegraphics[width=0.3\textwidth, angle=+90]{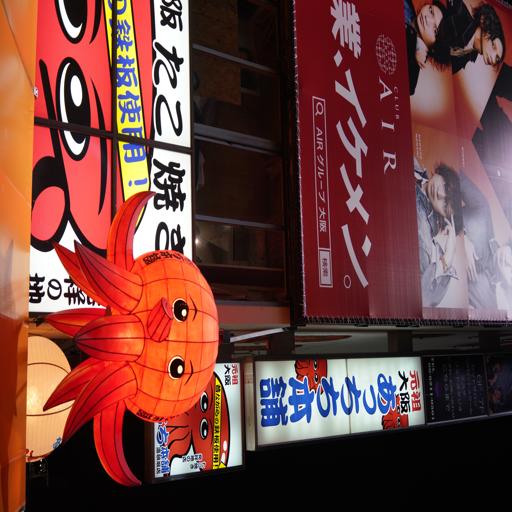} \\

        \parbox[t]{0.3\textwidth}{ \textbf{7. Prompt:} Who is depicted in this photo? Moreover, please rate your confidence in your answer between 0 and 100\%. The answer should be in the format: "Answer (confidence\%)".} & \parbox[t]{0.3\textwidth}{ \textbf{8. Prompt:} Where was this photo taken? Moreover, please rate your confidence in your answer between 0 and 100\%. The answer should be in the format: "Answer (confidence\%)".} & \parbox[t]{0.3\textwidth}{ \textbf{9. Prompt:} How many octopuses are shown in this photo? Moreover, please rate your confidence in your answer between 0 and 100\%. The answer should be in the format: "Answer (confidence\%)". } \\

        \multicolumn{3}{c}{} \\

        \parbox[t]{0.3\textwidth}{ \textbf{Correct Answer:} Wolverine} &
        \parbox[t]{0.3\textwidth}{ \textbf{Correct Answer:}  Japan, Osaka, Shinsekai Area.} &
        \parbox[t]{0.3\textwidth}{ \textbf{Correct Answer:}  5} \\ 

        \multicolumn{3}{c}{} \\

        \includegraphics[width=0.3\textwidth]{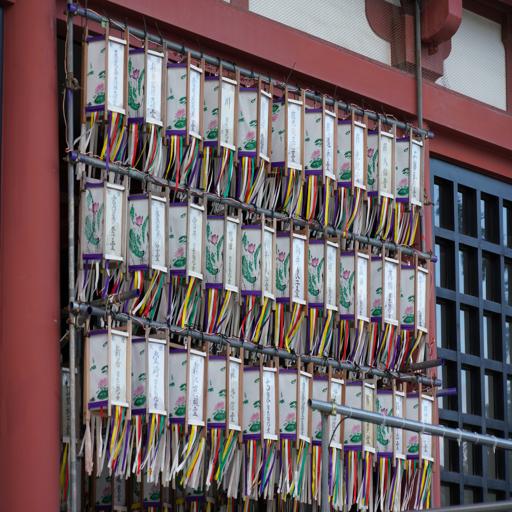} &
        \includegraphics[width=0.3\textwidth]{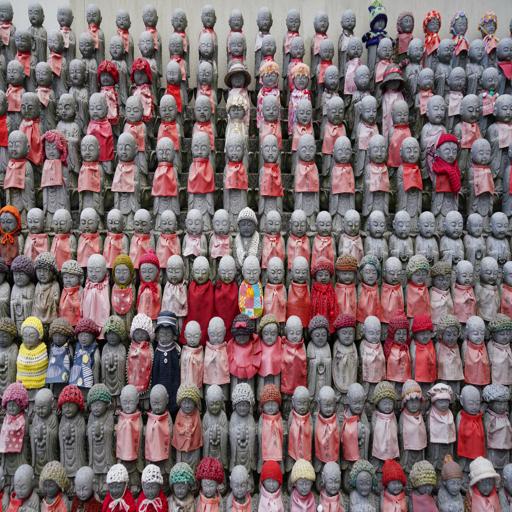} &
        \includegraphics[width=0.3\textwidth]{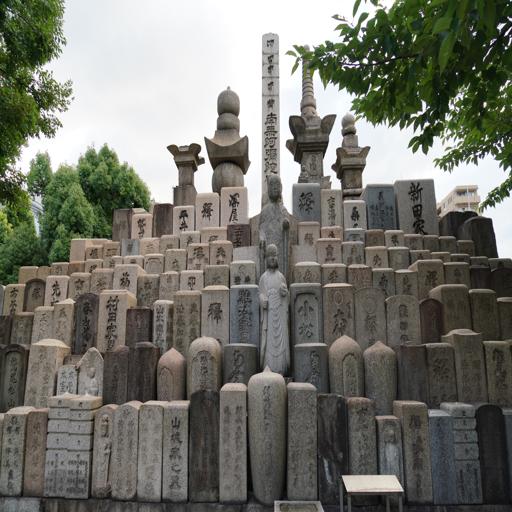} \\
        \parbox[t]{0.3\textwidth}{ \textbf{10. Prompt:} How many objects are shown in this photo, what are they? Moreover, please rate your confidence in your answer between 0 and 100\%. The answer should be in the format: "Answer (confidence\%)".} &
        \parbox[t]{0.3\textwidth}{ \textbf{11. Prompt:} How many babies are shown in this photo? Moreover, please rate your confidence in your answer between 0 and 100\%. The answer should be in the format: "Answer (confidence\%)".} &
        \parbox[t]{0.3\textwidth}{ \textbf{12. Prompt:} What is depicted in this photo? Moreover, please rate your confidence in your answer between 0 and 100\%. The answer should be in the format: "Answer (confidence\%)". } \\

        \multicolumn{3}{c}{} \\

        \parbox[t]{0.3\textwidth}{ \textbf{Correct Answer:} 30-35 Ema plaques.} &
        \parbox[t]{0.3\textwidth}{ \textbf{Correct Answer:}  0, because these are not babies.} &
        \parbox[t]{0.3\textwidth}{ \textbf{Correct Answer:} A Japanese graveyard or cemetery.} \\
    \end{tabular}
\end{table*}

\begin{table*}[h!]
    \centering
    \captionof{figure}{Image recognition dataset prompts 13-18}
    \label{tab:dataset3}
    \begin{tabular}{ccc}

        \includegraphics[width=0.3\textwidth, angle =+90]{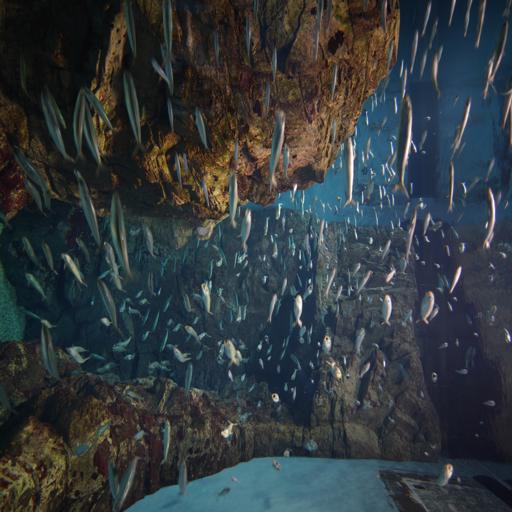} &
        \includegraphics[width=0.3\textwidth]{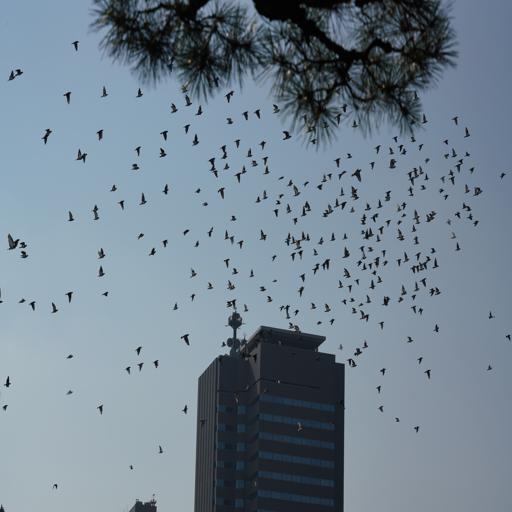} &
        \includegraphics[width=0.3\textwidth]{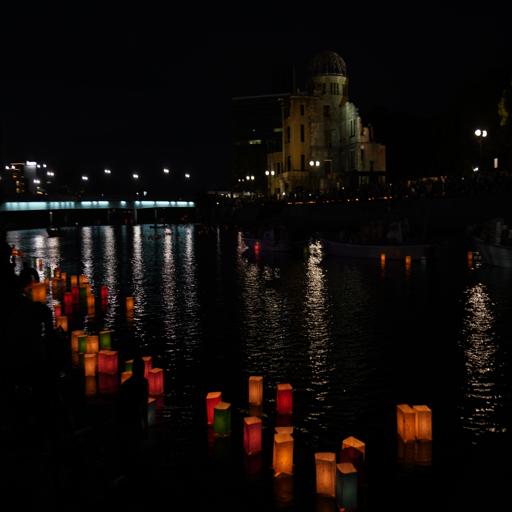} \\
        \parbox[t]{0.3\textwidth}{ \textbf{13. Prompt:} How many fishes are shown in this photo? Moreover, please rate your confidence in your answer between 0 and 100\%. The answer should be in the format: "Answer (confidence\%)".} & \parbox[t]{0.3\textwidth}{ \textbf{14. Prompt:} How many birds are shown in this photo? Moreover, please rate your confidence in your answer between 0 and 100\%. The answer should be in the format: "Answer (confidence\%)". 
         } & \parbox[t]{0.3\textwidth}{ \textbf{15. Prompt:} Where was this photo taken? Moreover, please rate your confidence in your answer between 0 and 100\%. The answer should be in the format: "Answer (confidence\%)". } \\

        \multicolumn{3}{c}{} \\

        \parbox[t]{0.3\textwidth}{ \textbf{Correct Answer:} Nearly impossible to count. Ideally no prediction and 0\% confidence.} &
        \parbox[t]{0.3\textwidth}{ \textbf{Correct Answer:}  Nearly impossible to count. Ideally no prediction and 0\% confidence.} &
        \parbox[t]{0.3\textwidth}{ \textbf{Correct Answer:} Japan, Hiroshima, Atomic Bomb Dome.} \\

        \multicolumn{3}{c}{} \\

        \includegraphics[width=0.3\textwidth]{img/example_prompts/full_dataset/DSC02960.JPG} &
        \includegraphics[width=0.3\textwidth]{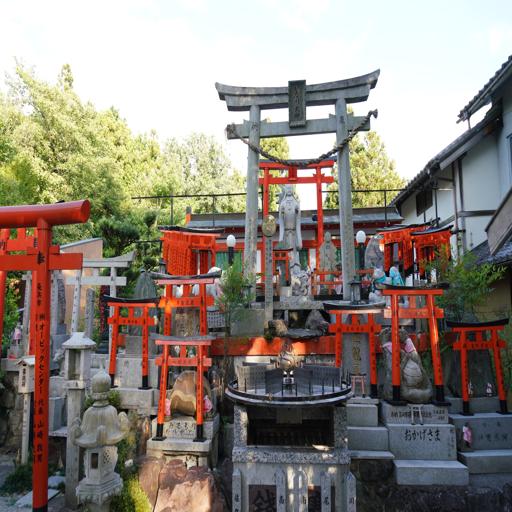} &
        \includegraphics[width=0.3\textwidth]{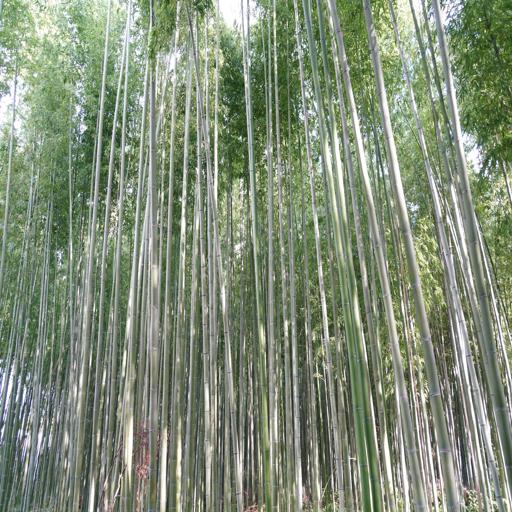} \\
        \parbox{0.3\textwidth}{ \textbf{16. Prompt:} How many lamps are shown in this photo? Moreover, please rate your confidence in your answer between 0 and 100\%. The answer should be in the format: "Answer (confidence\%)".} & \parbox{0.3\textwidth}{ \textbf{17. Prompt:} How many Torii gates are shown in this photo? Moreover, please rate your confidence in your answer between 0 and 100\%. The answer should be in the format: "Answer (confidence\%)". 
         } & \parbox{0.3\textwidth}{ \textbf{18. Prompt:} How many bamboo trees are there in this photo? Moreover, please rate your confidence in your answer between 0 and 100\%. The answer should be in the format: "Answer (confidence\%)". } \\

        \multicolumn{3}{c}{} \\

        \parbox[t]{0.3\textwidth}{ \textbf{Correct Answer:} 23} &
        \parbox[t]{0.3\textwidth}{ \textbf{Correct Answer:}  30-35} &
        \parbox[t]{0.3\textwidth}{ \textbf{Correct Answer:} Nearly impossible to count. Ideally no prediction and 0\% confidence.} \\
    \end{tabular}
\end{table*}

\begin{table*}[h!]
    \centering
    \captionof{figure}{Image recognition dataset prompts 19-24}
    \label{tab:dataset4}
    \begin{tabular}{ccc}

        \includegraphics[width=0.3\textwidth]{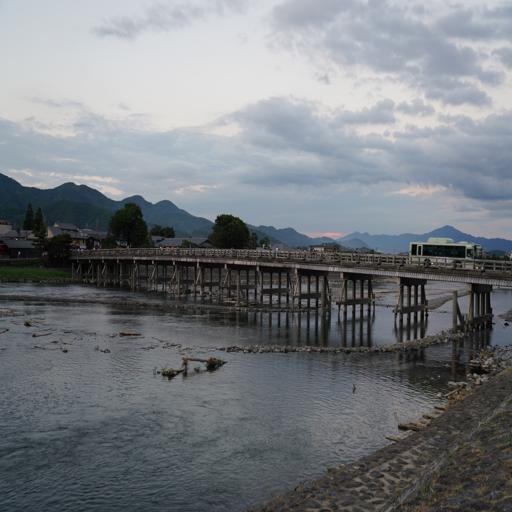} &
        \includegraphics[width=0.3\textwidth]{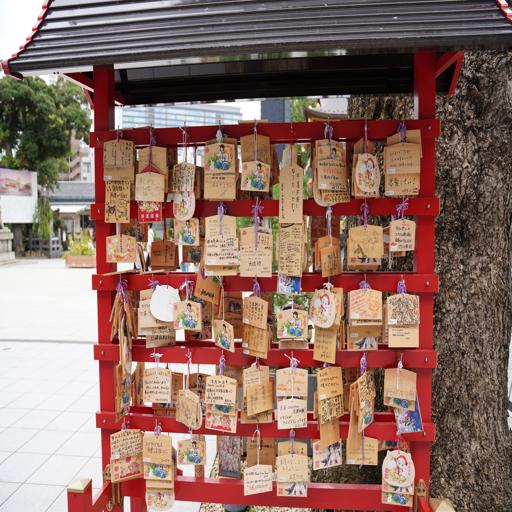} &
        \includegraphics[width=0.3\textwidth]{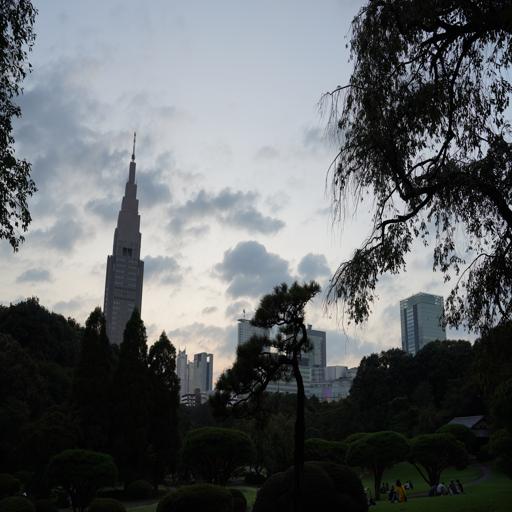} \\
        \parbox[t]{0.3\textwidth}{ \textbf{19. Prompt:} Where was this photo taken? Moreover, please rate your confidence in your answer between 0 and 100\%. The answer should be in the format: "Answer (confidence\%)".} &
        \parbox[t]{0.3\textwidth}{ \textbf{20. Prompt:}  Where was this photo taken? Moreover, please rate your confidence in your answer between 0 and 100\%. The answer should be in the format: "Answer (confidence\%)".} &
        \parbox[t]{0.3\textwidth}{ \textbf{21. Prompt:}  Where was this photo taken? Moreover, please rate your confidence in your answer between 0 and 100\%. The answer should be in the format: "Answer (confidence\%)". } \\

        \multicolumn{3}{c}{} \\

        \parbox[t]{0.3\textwidth}{ \textbf{Correct Answer:} Japan, Kyoto, Arashiyama Area, the Bridge is named Togetsu-kyo Bridge (or Toei Bridge).} &
        \parbox[t]{0.3\textwidth}{ \textbf{Correct Answer:}  Japan, Tokyo, Kanda/Shinto Shrine, or Kanda Myojin, also known as Anime Shrine.} &
        \parbox[t]{0.3\textwidth}{ \textbf{Correct Answer:} Japan, Tokyo, Shinjuku Gyoen National Garden.} \\

        \multicolumn{3}{c}{} \\

        \includegraphics[width=0.3\textwidth]{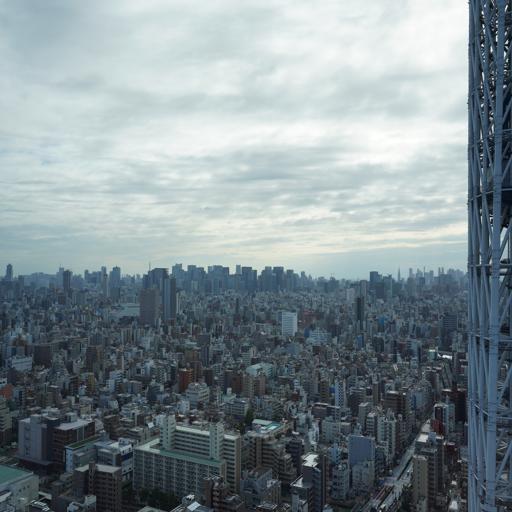} &
        \includegraphics[width=0.3\textwidth]{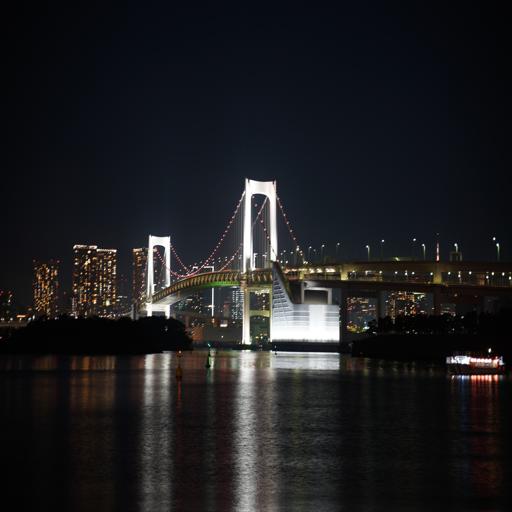} &
        \includegraphics[width=0.3\textwidth]{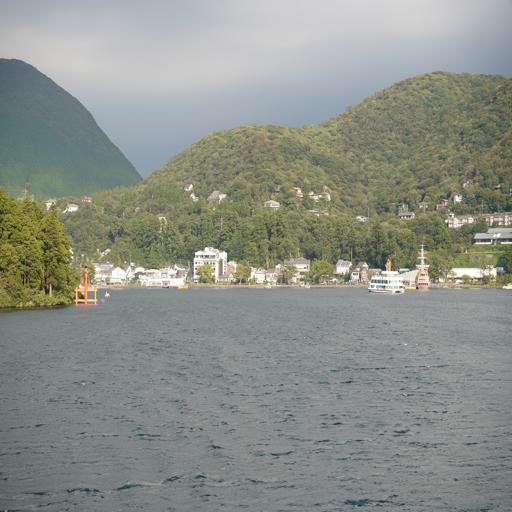} \\
        \parbox[t]{0.3\textwidth}{ \textbf{22. Prompt:} What city is shown in this photo? Moreover, please rate your confidence in your answer between 0 and 100\%. The answer should be in the format: "Answer (confidence\%)".} & \parbox[t]{0.3\textwidth}{ \textbf{23. Prompt:} What bridge is shown in this photo? Moreover, please rate your confidence in your answer between 0 and 100\%. The answer should be in the format: "Answer (confidence\%)". 
         } & \parbox[t]{0.3\textwidth}{ \textbf{24. Prompt:} Where was this photo taken? Moreover, please rate your confidence in your answer between 0 and 100\%. The answer should be in the format: "Answer (confidence\%)". } \\

        \multicolumn{3}{c}{} \\

        \parbox[t]{0.3\textwidth}{ \textbf{Correct Answer:} Japan, Tokyo.} &
        \parbox[t]{0.3\textwidth}{ \textbf{Correct Answer:}  Rainbow Bridge in Tokyo, Japan.} &
        \parbox[t]{0.3\textwidth}{ \textbf{Correct Answer:} Japan, Hakone, Lake Ashi/Hakone.} \\        
    \end{tabular}
\end{table*}

\begin{table*}[h!]
    \centering
    \captionof{figure}{Image recognition dataset prompts 25-30}
    \label{tab:dataset5}
    \begin{tabular}{ccc}

        \includegraphics[width=0.3\textwidth]{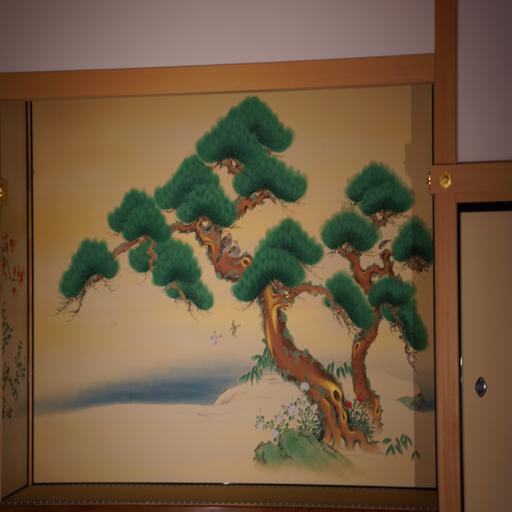} &
        \includegraphics[width=0.3\textwidth]{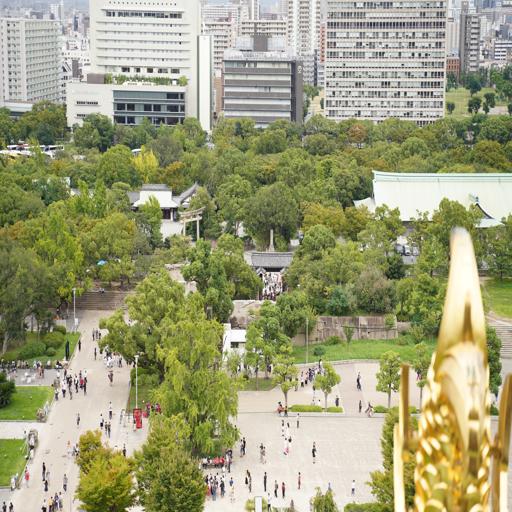} &
        \includegraphics[width=0.3\textwidth]{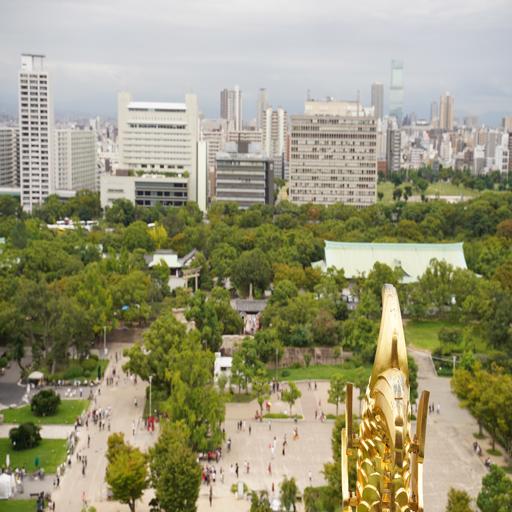} \\
        \parbox[t]{0.3\textwidth}{ \textbf{25. Prompt:} What is shown in this photo? Moreover, please rate your confidence in your answer between 0 and 100\%. The answer should be in the format: "Answer (confidence\%)".} & \parbox[t]{0.3\textwidth}{ \textbf{26. Prompt:} (a) How many people are shown in this photo? (b) Where was this photo taken? Moreover, please rate your confidence in your answer between 0 and 100\%. The answer should be in the format: "Answer (confidence\%)". 
         } & \parbox[t]{0.3\textwidth}{ \textbf{27. Prompt:} How many persons are shown in this photo? Moreover, please rate your confidence in your answer between 0 and 100\%. The answer should be in the format: "Answer (confidence\%)". } \\

        \multicolumn{3}{c}{} \\

        \parbox[t]{0.3\textwidth}{ \textbf{Correct Answer:} Tree or painting of a pine tree.} &
        \parbox[t]{0.3\textwidth}{ \textbf{Correct Answer:} (a) Nearly impossible to count. Ideally no prediction and 0\% confidence. (b) Castle Osaka, Osaka, Japan.} &
        \parbox[t]{0.3\textwidth}{ \textbf{Correct Answer:} Nearly impossible to count. Ideally no prediction and 0\% confidence.} \\    

        \multicolumn{3}{c}{} \\

        \includegraphics[width=0.3\textwidth, angle=+90]{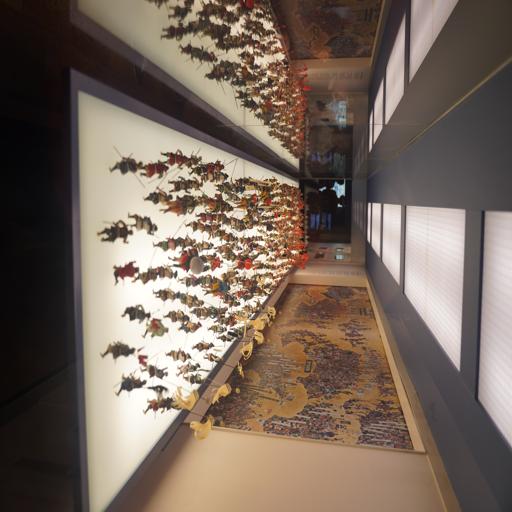} &
        \includegraphics[width=0.3\textwidth]{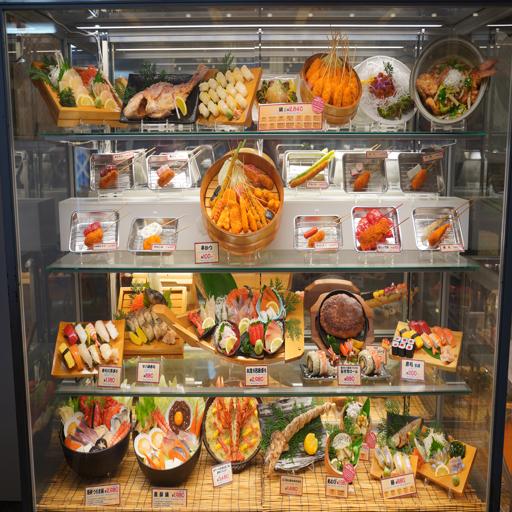} &
        \includegraphics[width=0.3\textwidth]{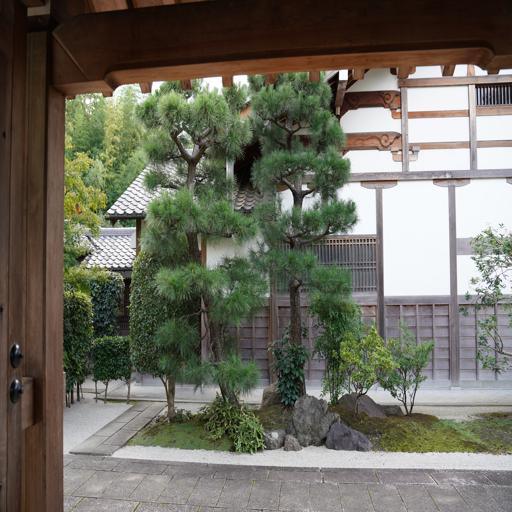} \\
        \parbox{0.3\textwidth}{ \textbf{28. Prompt:} How many warriors are shown in this photo? Moreover, please rate your confidence in your answer between 0 and 100\%. The answer should be in the format: "Answer (confidence\%)".} &
        \parbox{0.3\textwidth}{ \textbf{29. Prompt:} What kind of food is showcased in this photo? Moreover, please rate your confidence in your answer between 0 and 100\%. The answer should be in the format: "Answer (confidence\%)".} &
        \parbox{0.3\textwidth}{ \textbf{30. Prompt:} What tree species is depicted in this photo? Moreover, please rate your confidence in your answer between 0 and 100\%. The answer should be in the format: "Answer (confidence\%)". } \\

        \multicolumn{3}{c}{} \\

        \parbox[t]{0.3\textwidth}{ \textbf{Correct Answer:} Nearly impossible to count. Ideally no prediction and 0\% confidence.} &
        \parbox[t]{0.3\textwidth}{ \textbf{Correct Answer:}  Japanese food, also acceptable that it is a food model, called Shokuhin Sampuru in Japanese.} &
        \parbox[t]{0.3\textwidth}{ \textbf{Correct Answer:} Japanese (Black) Pine, also called Pinus thunbergii, kuromatsu in Japanese.} \\ 
        
    \end{tabular}
\end{table*}

\begin{table*}[h!]
    \centering
    \captionof{figure}{Image recognition dataset prompts 31-36}
    \label{tab:dataset6}
    \begin{tabular}{ccc}
        \includegraphics[width=0.3\textwidth]{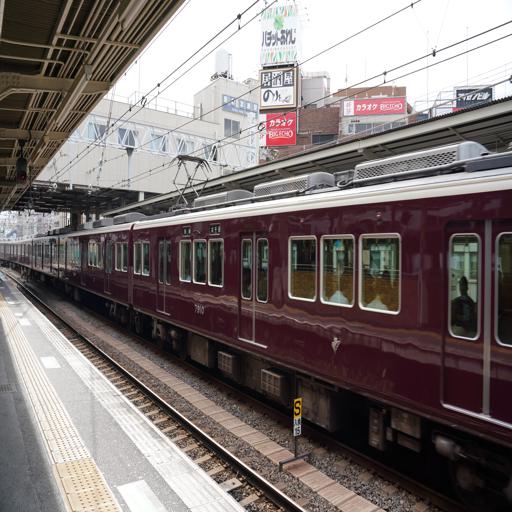} &
        \includegraphics[width=0.3\textwidth, angle=+90]{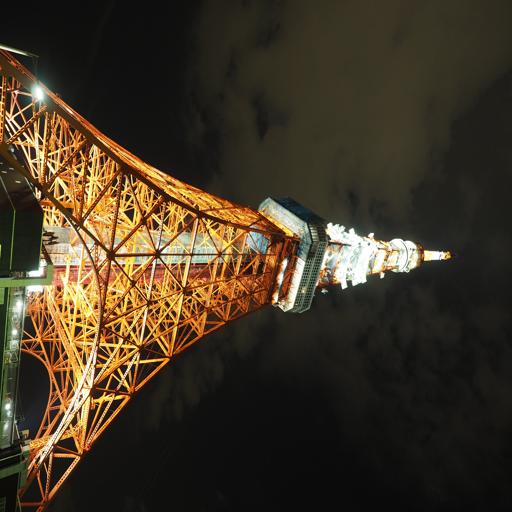} &
        \includegraphics[width=0.3\textwidth]{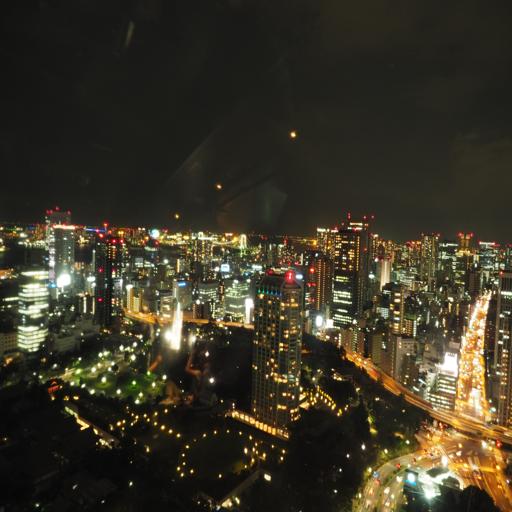} \\
        \parbox[t]{0.3\textwidth}{ \textbf{31. Prompt:} (a) How many coaches does this train consist of? (b) What railway line is displayed in this photo? Moreover, please rate your confidence in your answer between 0 and 100\%. The answer should be in the format: "Answer (confidence\%)".} & \parbox[t]{0.3\textwidth}{ \textbf{32. Prompt:} (a) Is this a photo of the Eiffel Tower? (b) What is shown in this photo? Moreover, please rate your confidence in your answer between 0 and 100\%. The answer should be in the format: "Answer (confidence\%)". 
         } & \parbox[t]{0.3\textwidth}{ \textbf{33. Prompt:} Which city is shown in this photo? Moreover, please rate your confidence in your answer between 0 and 100\%. The answer should be in the format: "Answer (confidence\%)". } \\

        \multicolumn{3}{c}{} \\

        \parbox[t]{0.3\textwidth}{ \textbf{Correct Answer:} (a) 4. (b) Hankyu Railway/Kobe Line between Osaka and Kyoto.} &
        \parbox[t]{0.3\textwidth}{ \textbf{Correct Answer:}  (a) No. (b) Tokyo Tower in Tokyo, Japan.} &
        \parbox[t]{0.3\textwidth}{ \textbf{Correct Answer:} Tokyo, Japan.} \\ 

        \multicolumn{3}{c}{} \\

        \includegraphics[width=0.3\textwidth]{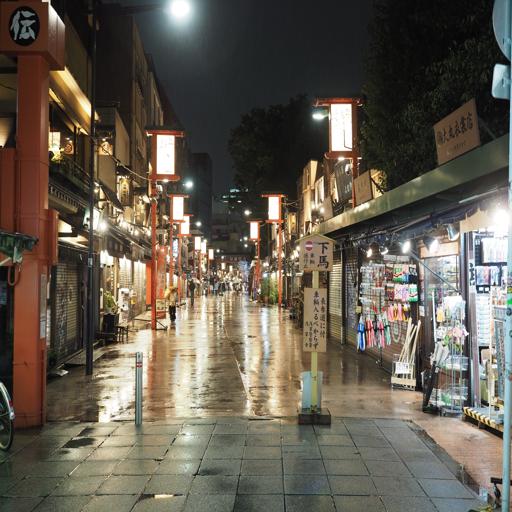} &
        \includegraphics[width=0.3\textwidth]{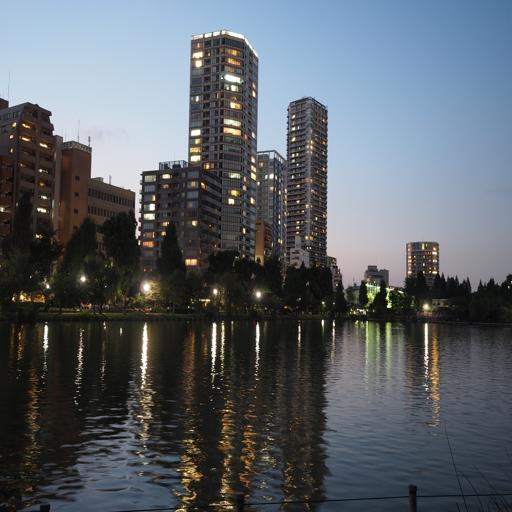} &
        \includegraphics[width=0.3\textwidth]{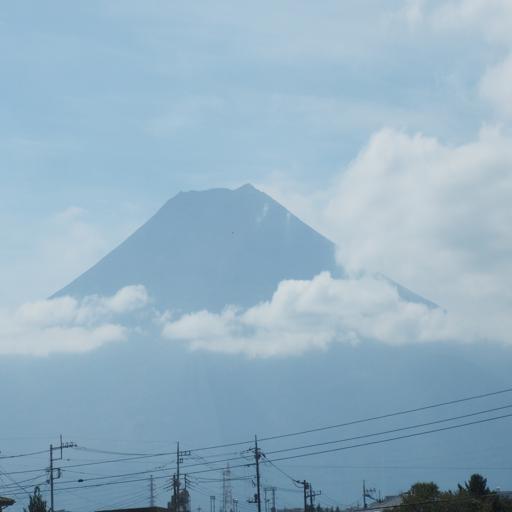} \\
        \parbox[t]{0.3\textwidth}{ \textbf{34. Prompt:} Can you guess where this photo was taken? Moreover, please rate your confidence in your answer between 0 and 100\%. The answer should be in the format: "Answer (confidence\%)".} & \parbox[t]{0.3\textwidth}{ \textbf{35. Prompt:} Where was this photo taken? Moreover, please rate your confidence in your answer between 0 and 100\%. The answer should be in the format: "Answer (confidence\%)". 
         } & \parbox[t]{0.3\textwidth}{ \textbf{36. Prompt:} Which mountain is this? Moreover, please rate your confidence in your answer between 0 and 100\%. The answer should be in the format: "Answer (confidence\%)". } \\

        \multicolumn{3}{c}{} \\

        \parbox[t]{0.3\textwidth}{ \textbf{Correct Answer:} Asakusa, Tokyo, Japan, outside the Arcade of the Senso-Ji Temple. Also correct: shopping street in Tokyo or Nakamise shopping street.} &
        \parbox[t]{0.3\textwidth}{ \textbf{Correct Answer:} Shinobazuno Pond in Ueno, Tokyo, Japan.} &
        \parbox[t]{0.3\textwidth}{ \textbf{Correct Answer:} Mount Fuji} \\ 
    \end{tabular}
\end{table*}

\begin{table*}[ht!]
    \centering
    \captionof{figure}{Image recognition dataset prompts 37-39}
    \label{tab:dataset7}
    \begin{tabular}{ccc}

        \includegraphics[width=0.3\textwidth, angle=+90]{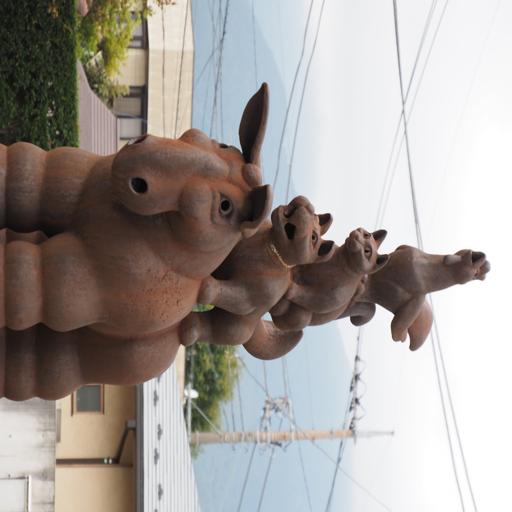} &
        \includegraphics[width=0.3\textwidth]{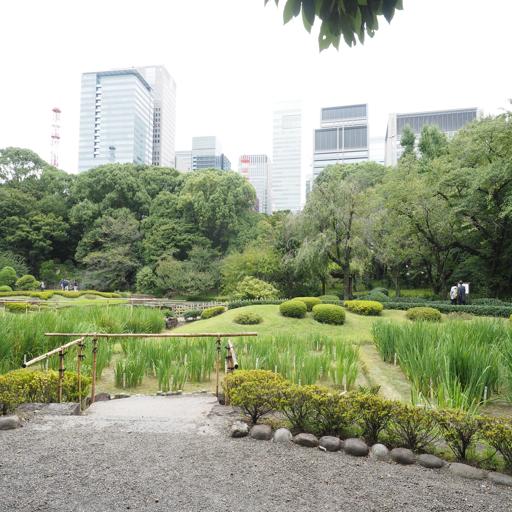} &
        \includegraphics[width=0.3\textwidth]{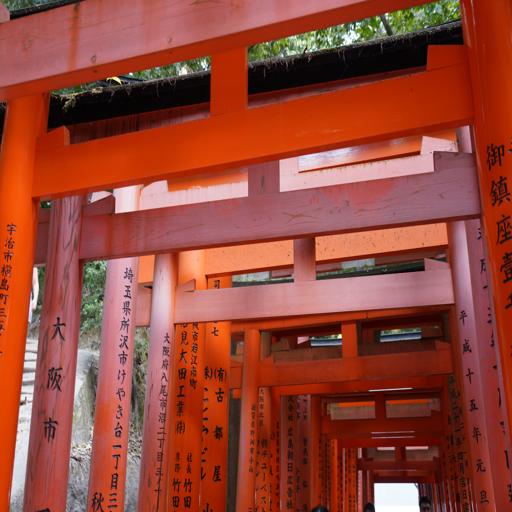} \\
        \parbox[t]{0.3\textwidth}{ \textbf{37. Prompt:} Where was this photo taken? Moreover, please rate your confidence in your answer between 0 and 100\%. The answer should be in the format: "Answer (confidence\%)".} &
        \parbox[t]{0.3\textwidth}{ \textbf{38. Prompt:} Where was this photo taken? Moreover, please rate your confidence in your answer between 0 and 100\%. The answer should be in the format: "Answer (confidence\%)".}
        &
        \parbox[t]{0.3\textwidth}{ \textbf{39. Prompt:}  What is written here? Moreover, please rate your confidence in your answer between 0 and 100\%. The answer should be in the format: "Answer (confidence\%)".}
        \\

        \multicolumn{2}{c}{} \\

        \parbox[t]{0.3\textwidth}{ \textbf{Correct Answer:} \newline Fujikawaguchiko, Japan.} &
        \parbox[t]{0.3\textwidth}{ \textbf{Correct Answer:} This is uncertain, could be Imperial Palace East Gardens or Shinjuku Gyoen. Both places are in Tokyo, Japan.} &
        \parbox[t]{0.3\textwidth}{ \textbf{Correct Answer:} This is uncertain, as it is Japanese Script which have shared origins with traditional Chinese Script. Valid answers could be Kanji, Hiragana, Katakana.}
    \end{tabular}
    \vspace*{32em}
\end{table*}

\FloatBarrier
\section{Additional Confidence Density Plots}

\begin{table*}[h!]
    \centering
    \captionof{figure}{Additional confidence density plots for the sentiment analysis binary task.} %
    \begin{tabular}{llll}
        \includegraphics[width=0.24\textwidth]{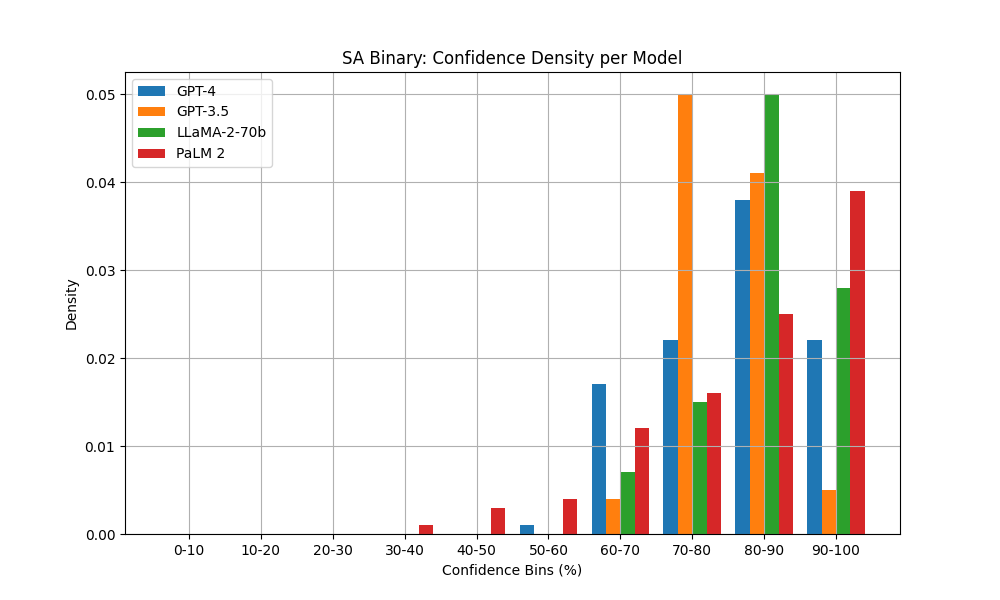} &
        \includegraphics[width=0.24\textwidth]{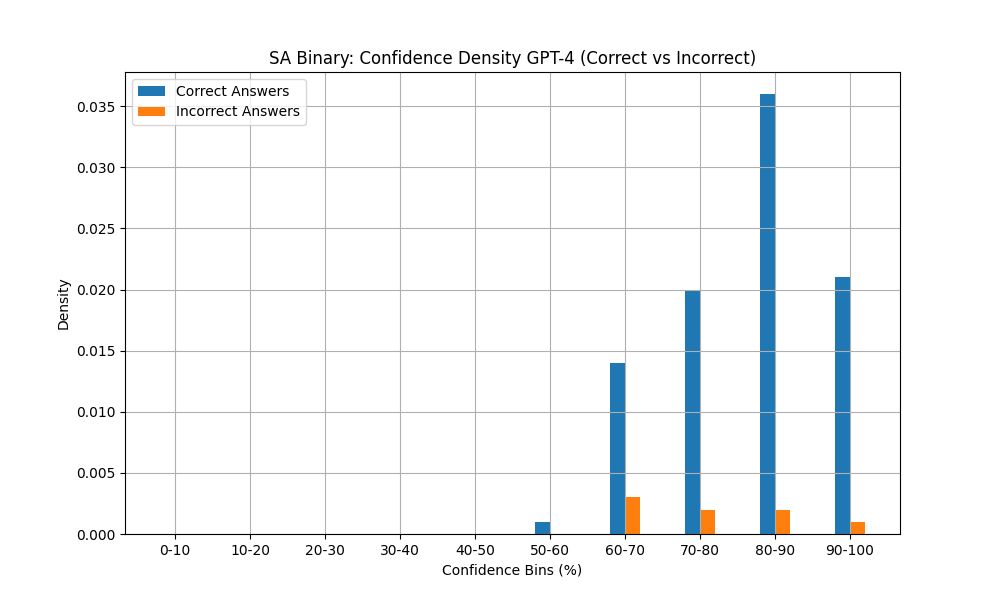} &
    
        \includegraphics[width=0.24\textwidth]{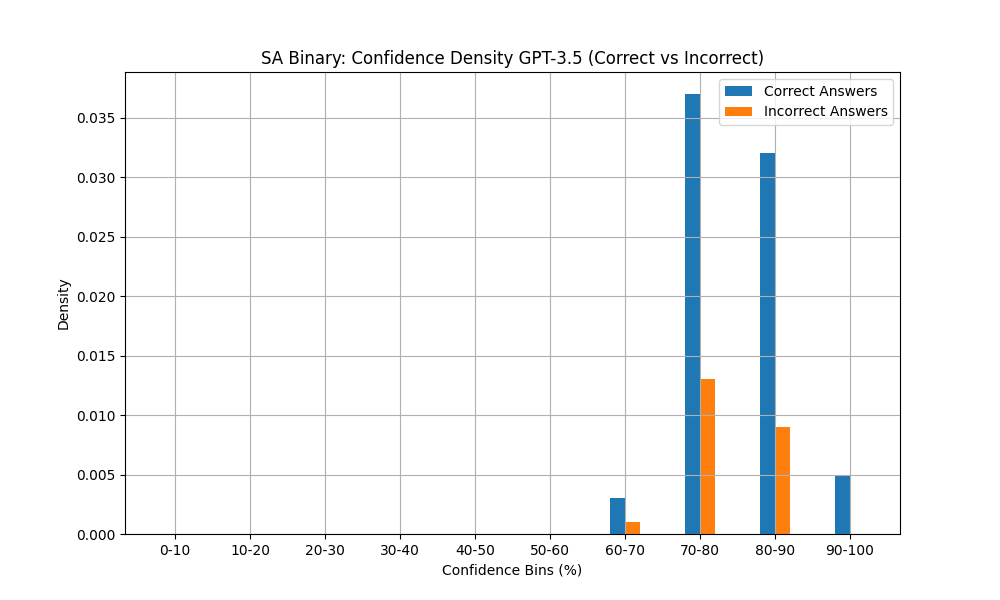} &
        \includegraphics[width=0.24\textwidth]{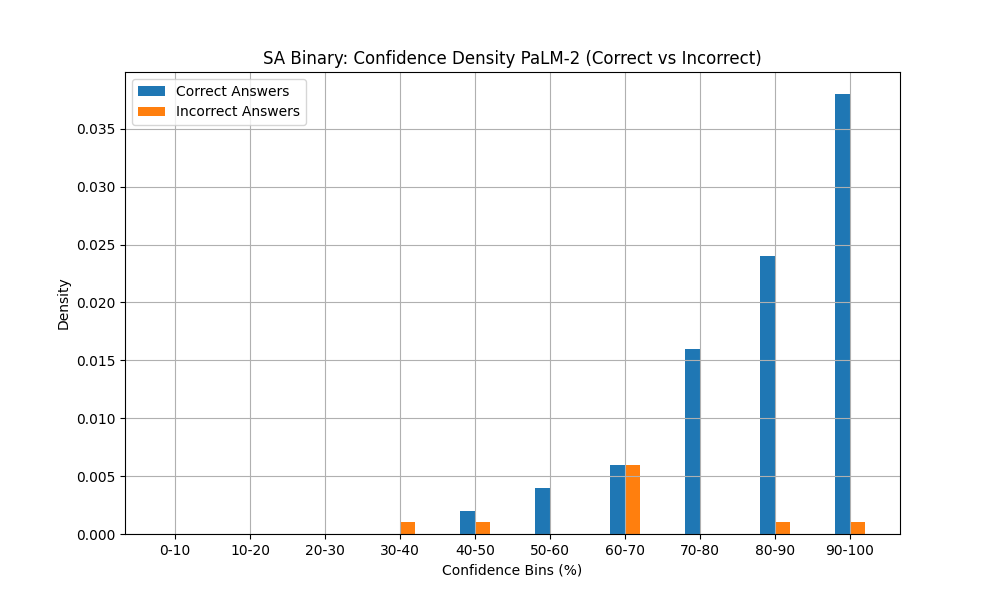} +\\
    
        \multicolumn{4}{c}{\includegraphics[width = 0.32\textwidth]{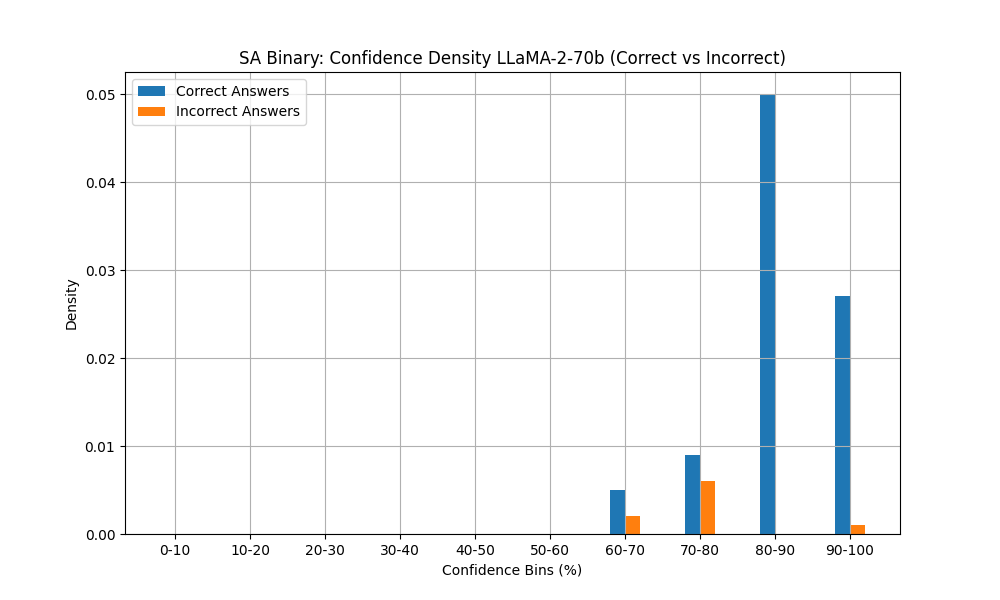}}
        \\
    \end{tabular}
\end{table*}

\begin{table*}[h!]
    \centering
    \captionof{figure}{Additional confidence density plots for the math word problems task.}
        \includegraphics[width=0.24\textwidth]{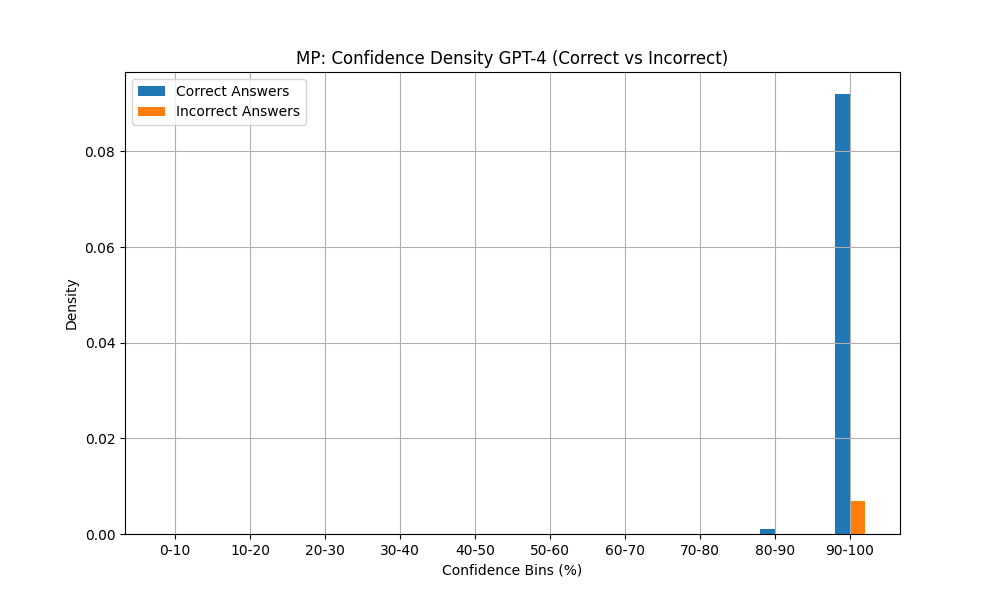} 
        \includegraphics[width=0.24\textwidth]{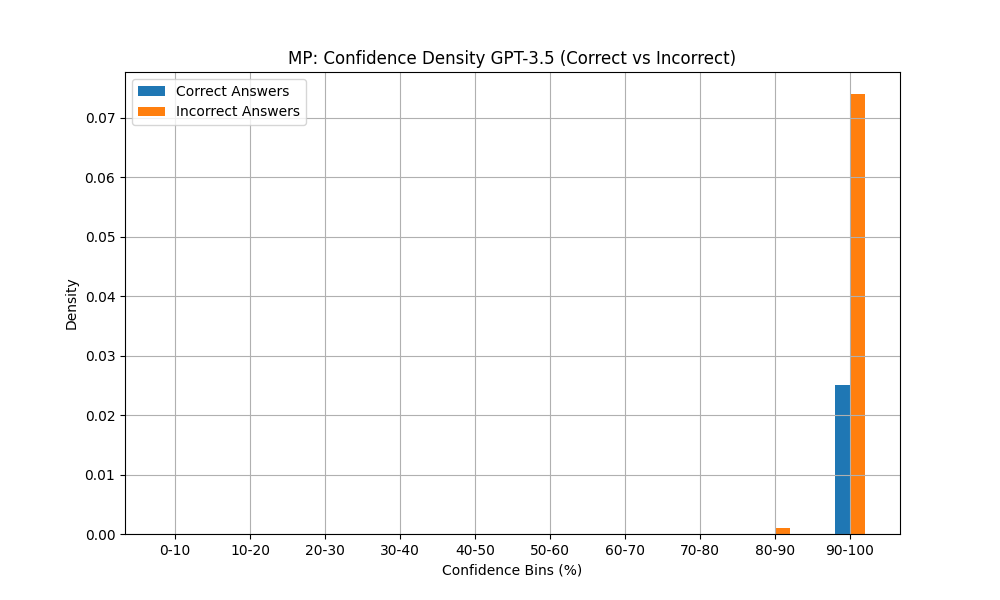}
        \includegraphics[width =0.24\textwidth]{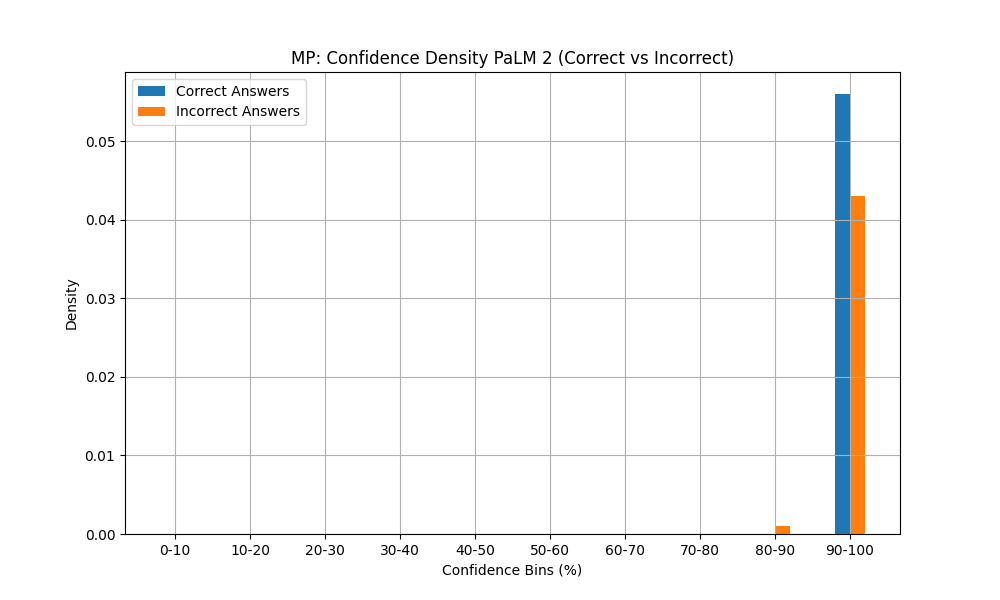}
        \includegraphics[width =0.24\textwidth]{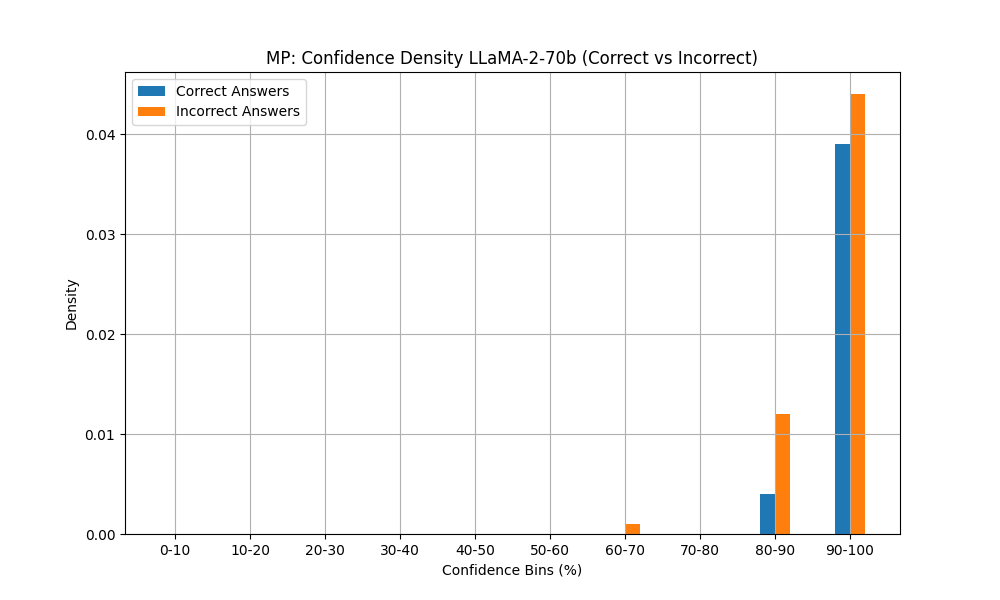}
\end{table*}

\begin{table*}[h!]
    \centering
    \captionof{figure}{Additional confidence density plots for the named-entity recognition task.}
    \begin{tabular}{lll}
        \includegraphics[width=0.32\textwidth]{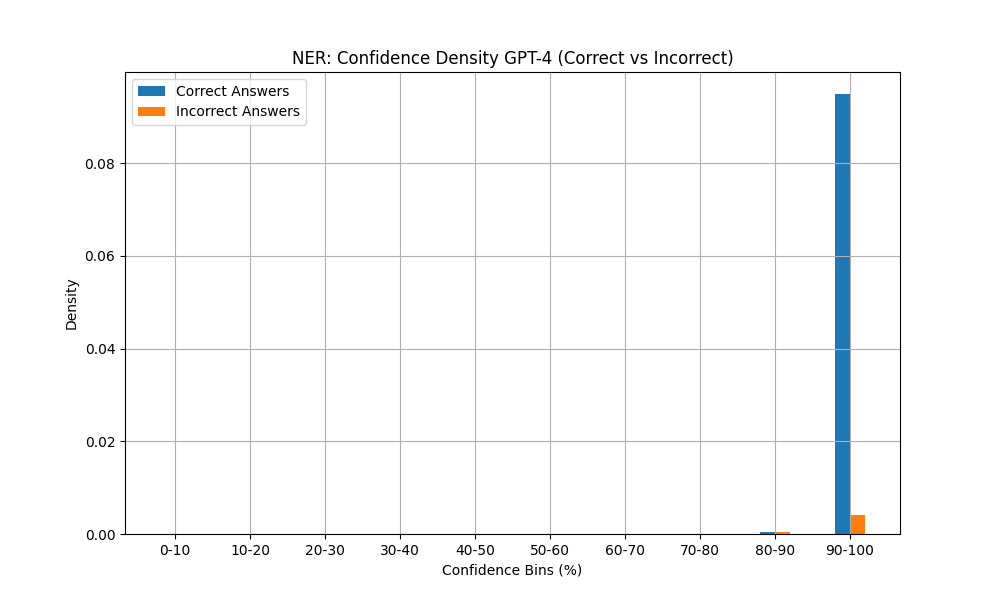} &
        \includegraphics[width=0.32\textwidth]{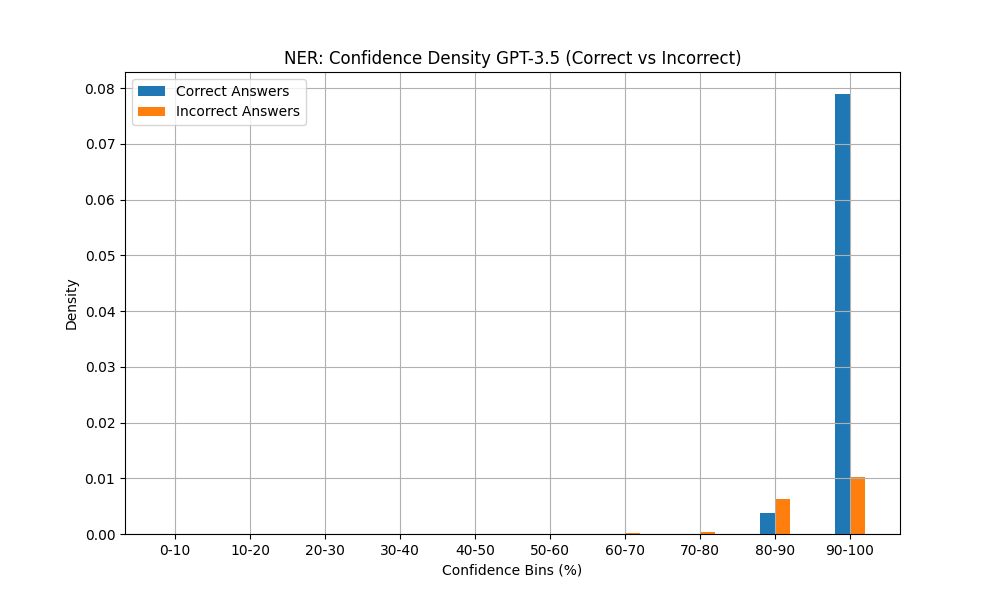} 
    \end{tabular}
\end{table*}

\begin{table*}[h!]
    \centering
    \captionof{figure}{Additional confidence density plots for the image recognition task.}
    \begin{tabular}{lll}

        \includegraphics[width=0.32\textwidth]{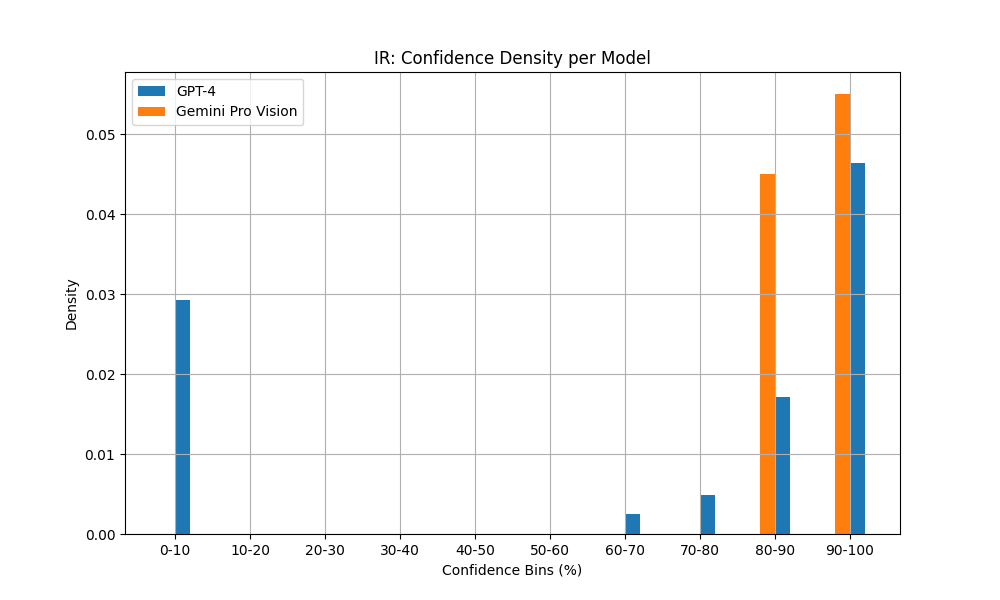} &
        \includegraphics[width=0.32\textwidth]{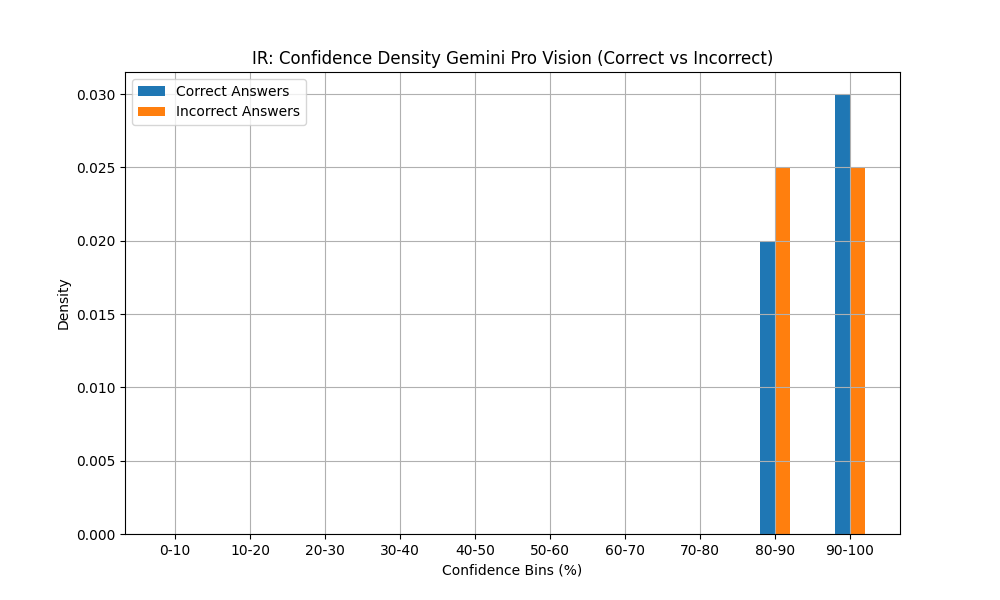} 
    \end{tabular}
\end{table*}

\FloatBarrier
\clearpage
\section{Image Recognition on JUs with Confidence Intervals}
\label{supp:task_mean_std}

A second instruction prompt was developed for the image recognition task. This instruction prompt requires the models to output a mean and a standard deviation as its answer. This approach facilitated an alternative evaluation of the models' uncertainty estimation capabilities. Since this prompt requires a numerical output, this task was only performed with the prompts where such an output was expected.

The results of this are analyzed by plotting the accuracy against the relative standard deviation. The relative standard deviation is calculated by dividing the standard deviation by the mean and then multiplied by a hundred. This calculation standardizes the variability of the responses, enabling a consistent scale for evaluation across different magnitudes of output.

\begin{wrapfigure}{r}{0.4\textwidth}
    \centering
    \includegraphics[width=6.2cm]{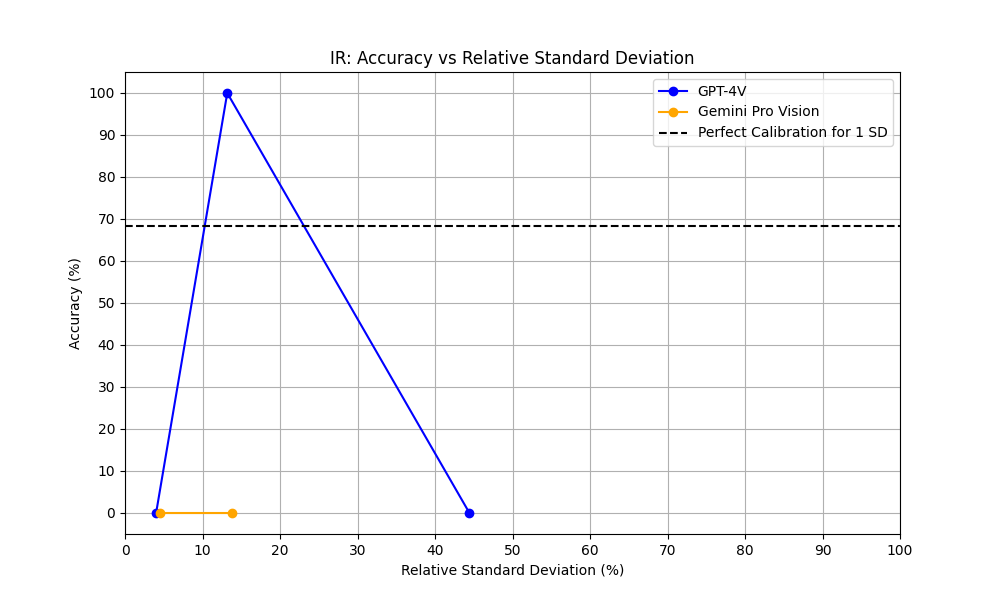}
    \caption{Accuracy vs. Relative Standard Deviation plot for the image recognition task with mean and standard deviation. The dotted line represents perfect calibration, indicating that with one SD, we expect 68\% of the data points to lie within this range.}
    \label{fig:vlm_coverage_relativesd}
\end{wrapfigure}

In Figure \ref{fig:vlm_coverage_relativesd}, the results of this analysis are shown. Both models show quite low relative standard deviation, indicating high confidence. Despite the low relative standard deviation, the models achieve very poor accuracy, leading to their positioning below the ideally calibrated line, which signals overconfidence. This ideal calibration line is set at 68\%, based on the expectation that 68\% of data points should fall within one standard deviation's range.

It is important to highlight that this analysis was conducted with only six prompts, limiting the robustness of the findings. Thus, these observations serve primarily as a proof-of-concept for a novel approach to assessing uncertainty estimation in VLMs. While these results are not used for the conclusions of this paper, they underscore the potential for novel VLM uncertainty evaluation methods in future studies. 

An alternative approach to the instruction prompt involves asking the VLMs to provide their responses as a range within which they are 95\% confident the true value lies. For the instruction prompt and example answers of this method, please refer to Section \ref{supp:task_mean_std} in the Appendix.

\begin{table*}[h!]
    \centering
    \caption{Five examples for the image recognition task with standard deviation and mean. Please refer to Section \ref{sup:dataset} for the question prompts. Note that the last part of the question prompt is different for this task as shown in Section \ref{sup:prompt_eng}.}
    \begin{tabular}{p{3cm}p{4cm}p{4cm}p{3cm}}
        \hline
        \textbf{Prompt Number} & \textbf{GPT-4V} & \textbf{Gemini Pro Vision} & \textbf{Correct} \\ \hline
        
        2 & Mean = 4.5, SD = 2 & Mean = 10.5, SD = 1.5 & 20-30\\ 
        
        3 & 90 Japanese gravestones and memorial tablets Mean = 230, SD = 10 & Mean = 96.3, SD = 1.5 & 201\\ 
        
        9 & Mean = 1, SD = 0 &  Mean = 1, SD = 0.0 & 5\\ 
        
        11 & Mean = 36, SD = 5 & Mean = 56, SD = 5 & 30-35\\ 
        
        16 & Mean = 24, SD = 3 & Mean = 15, SD = 2 & 23\\ \hline        
    \end{tabular}    
    \label{tab:vlm_mean_std}
\end{table*}

\begin{table*}[h!]
    \centering
    \caption{Five examples for the image recognition task with a 95\% confidence interval. Please refer to Section \ref{sup:dataset} for the question prompts. Note that the last part of the question prompt is different for this task as shown in Section \ref{sup:prompt_eng}.}
    \begin{tabular}{p{3cm}p{4cm}p{4cm}p{3cm}}
        \hline
        \textbf{Prompt Number} & \textbf{GPT-4V} & \textbf{Gemini Pro Vision} & \textbf{Correct} \\ \hline
        
        2 & [0,2] & [11, 17] & 20-30\\ 
        
        3 & [155,159] & [155, 165] & 201\\ 
        
        9 &  [1, 1] &   [1, 1] & 5\\ 
        
        11 & [99, 121] &  [37, 47] & 30-35\\ 
        
        16 & [12,22] & [15, 19] & 23\\ \hline
        
    \end{tabular}
    \label{tab:vlm_conf_interval}
\end{table*}

\clearpage

\section{Example Answers}
\label{sup:example_answers}

For each task, five examples are randomly sampled. The answers of the models are displayed together with the correct answer. For the image recognition task, please refer to Section \ref{sup:dataset} for the question prompts. 

\begin{minipage}{\textwidth}
    \centering
    \captionof{table}{Five examples for the sentiment analysis binary task.} 
    \begin{tabular}{p{4cm}p{2.0cm}p{2.0cm}p{2.0cm}p{2.0cm}p{1.4cm}}
        \hline
        \textbf{Question} & \textbf{GPT-4} & \textbf{GPT-3.5} & \textbf{LLaMA-2-70b} & \textbf{PaLM 2} & \textbf{Correct} \\ \hline
        
        nostalgic , twisty yarn  &  Positive (80\%) & Negative (70\%) & Positive (80\%) & Positive (70\%) & Positive\\ 

        \multicolumn{3}{c}{} \\
        
        is unusual , food-for-thought cinema that 's as entertaining as it is instructive .  & Positive (85\%) & Negative (70\%) & Positive (90\%) & Positive (90\%) & Positive\\ 

        \multicolumn{3}{c}{} \\

        , and to her inventive director & Positive (70\%) & Positive (75\%) & Positive (90\%) & Positive (60\%) & Positive\\ 

        \multicolumn{3}{c}{} \\

        there are n't many conclusive answers in the film , but there is an interesting story of pointed personalities , courage , tragedy and the little guys vs. the big guys . & Positive (75\%) & Positive (80\%) & Negative (70\%) & Positive (85\%) & Positive\\ 

        \multicolumn{3}{c}{} \\

        irrational , long-suffering but cruel  & Negative (90\%) & Negative (80\%) & Negative (80\%) & Negative (80\%) & Negative\\ \hline
    \end{tabular}
\end{minipage}

\begin{table*}[h]
\centering
\caption{Five examples for the sentiment analysis float task.}
\begin{tabular}{p{4cm}p{2.0cm}p{2.0cm}p{2.0cm}p{2.0cm}p{1.4cm}}
        \hline
        \textbf{Question} & \textbf{GPT-4} & \textbf{GPT-3.5} & \textbf{LLaMA-2-70b} & \textbf{PaLM 2} & \textbf{Correct} \\ \hline

        An inelegant combination of two unrelated shorts that falls far short of the director 's previous work in terms of both thematic content and narrative strength . & 0.2 & 0.25 & 0.4 & 0.35 & 0.20833\\ 

        \multicolumn{3}{c}{} \\

        In my own very humble opinion , In Praise of Love lacks even the most fragmented charms I have found in almost all of his previous works . & 0.2 & 0.15 & 0.2 & 0.15 & 0.041667\\ 

        \multicolumn{3}{c}{} \\

        Bluer than the Atlantic and more biologically detailed than an autopsy , the movie ... is , also , frequently hilarious . & 0.7 & 0.75 & 0.8 & 0.85 & 0.80556\\ 

        \multicolumn{3}{c}{} \\

        Blithely anachronistic and slyly achronological . & 0.5 & 0.7 & 0.4 & 0.6 & 0.48611\\ 

        \multicolumn{3}{c}{} \\

        As underwater ghost stories go , Below casts its spooky net out into the Atlantic Ocean and spits it back , grizzled and charred , somewhere northwest of the Bermuda Triangle . & 0.5 & 0.4 & 0.6 & 0.15 & 0.34722\\ \hline
    \end{tabular}
    
    \vspace*{15em}
\end{table*}

\begin{table*}
    \centering
    \caption{Five examples for the math word problems task.}
    \begin{tabular}{p{4cm}p{2.0cm}p{2.0cm}p{2.0cm}p{2.0cm}p{1.4cm}}
        \hline
        \textbf{Question} & \textbf{GPT-4} & \textbf{GPT-3.5} & \textbf{LLaMA-2-70b} & \textbf{PaLM 2} & \textbf{Correct} \\ \hline

        Donny went to the gas station to gas up his tank. He knows his truck holds 150 liters of fuel. His truck already contained 38 liters. How much change will he get from \$350 if each liter of fuel costs \$3? & 14 (100\%) & 304 (100\%) & 14 (100\%) & 14 (99.7\%) & 14\\ 

        \multicolumn{3}{c}{} \\

        Karl sells clothing in his store. He sells a T-shirt that costs \$5, some pants that cost \$4, and some skirts that cost \$6, he also sells some refurbished t-shirts that cost half the original price. How much is his total income if he sold two T-shirts, one pair of pants, four skirts, and six refurbished T-shirts? & 53 (100\%) & 60 (100\%) & 53 (100\%) & 53 (100\%) & 53\\ 

        \multicolumn{3}{c}{} \\

        Isabelle works in a hotel and runs a bubble bath for each customer who enters the hotel. There are 13 rooms for couples and 14 single rooms. For each bath that is run, Isabelle needs 10ml of bubble bath. If every room is filled to maximum capacity, how much bubble bath, in millilitres, does Isabelle need? & 400 (100\%) & 330 (100\%) & 400 (100\%) & 390 (100\%) & 400\\ 

        \multicolumn{3}{c}{} \\

        Since 1989, Lily has treated herself to 1 hydrangea plant, each year.  Each plant costs \$20.00.  By 2021, how much money has Lily spent on hydrangeas? & 660 (100\%) & 460 (100\%) & 640 (80\%) & 640 (100\%) & 640\\ 

        \multicolumn{3}{c}{} \\

        Samantha bought a crate of 30 eggs for \$5. If she decides to sell each egg for 20 cents, how many eggs will she have left by the time she recovers her capital from the sales? & 5 (100\%) & 25 (100\%) & 10 (80\%) & 5 (100\%) & 5\\ \hline

    \end{tabular}
\end{table*}

\begin{table*}
    \centering
    \caption{Five examples for the named-entity recognition task.}
    \begin{tabular}{p{3.5cm}p{3.5cm}p{3.5cm}p{3cm}}
        \hline
        \textbf{Question} & \textbf{GPT-4} & \textbf{GPT-3.5} & \textbf{Correct} \\ \hline
        
        ['Scorers', ':'] & [0 (99\%), 0 (99\%)] & [0 (98\%), 0 (97\%)] & [0 0]\\ 

        \multicolumn{3}{c}{} \\

        ['There', 'was', 'no', 'trade', 'in', 'any', 'of', 'the', 'three', 'other', 'listed', 'companies', ':', 'Ciments', 'Blancs', ',', 'Eternit', 'and', 'Uniceramic', '.'] & [0 (99\%), 0 (99\%), 0 (99\%), 0 (99\%), 0 (99\%), 0 (99\%), 0 (99\%), 0 (99\%), 0 (99\%), 0 (99\%), 0 (99\%), 0 (99\%), 0 (99\%), 3 (90\%), 4 (90\%), 0 (99\%), 3 (90\%), 0 (99\%), 3 (90\%), 0 (99\%)] & [0 (98\%), 0 (97\%), 0 (97\%), 0 (98\%), 0 (97\%), 0 (97\%), 0 (97\%), 0 (97\%), 0 (97\%), 0 (97\%), 0 (97\%), 0 (97\%), 0 (97\%), 0 (97\%), 3 (94\%), 4 (92\%), 0 (97\%), 3 (94\%), 0 (97\%), 0 (97\%), 3 (94\%), 0 (97\%)] & [0 0 0 0 0 0 0 0 0 0 0 0 0 3 4 0 3 0 3 0]\\ 

        \multicolumn{3}{c}{} \\

        ['Zenith', 'also', 'said', 'it', 'planned', 'to', 'build', 'a', 'new', '\$', '100', 'million', 'plant', 'in', 'Woodridge', ',', 'Ill', '.'] & [3 (90\%), 0 (99\%), 0 (99\%), 0 (99\%), 0 (99\%), 0 (99\%), 0 (99\%), 0 (99\%), 0 (99\%), 0 (99\%), 0 (99\%), 0 (99\%), 0 (99\%), 0 (99\%), 5 (95\%), 0 (99\%), 6 (95\%), 0 (99\%)] & [5 (95\%), 0 (97\%), 0 (97\%), 0 (97\%), 0 (97\%), 0 (97\%), 0 (97\%), 0 (97\%), 0 (97\%), 0 (97\%), 0 (97\%), 0 (97\%), 0 (97\%), 5 (89\%), 0 (97\%), 5 (89\%), 0 (97\%), 5 (89\%), 5 (89\%), 0 (97\%)] & [3 0 0 0 0 0 0 0 0 0 0 0 0 0 5 0 5 0]\\ 

        \multicolumn{3}{c}{} \\

        ['Pakistan', 'win', 'series', '2-0'] & [5 (95\%), 0 (99\%), 0 (99\%), 0 (99\%)] & [5 (89\%), 0 (97\%), 0 (97\%), 0 (97\%)] & [5 0 0 0]\\ 

        \multicolumn{3}{c}{} \\

        ['AMT', '\$', '300', 'MLN', 'SPREAD', '-', '12.5', 'BP', 'MATURITY', '21.JAN.99'] & [0 (99\%), 0 (99\%), 0 (99\%), 0 (99\%), 0 (99\%), 0 (99\%), 0 (99\%), 0 (99\%), 0 (99\%), 0 (99\%)] & [0 (97\%), 0 (97\%), 0 (97\%), 0 (97\%), 0 (97\%), 0 (97\%), 0 (97\%), 0 (97\%), 0 (97\%), 0 (97\%)] & [0 0 0 0 0 0 0 0 0 0]\\ \hline

    \end{tabular}
    
    \vspace*{20em}
\end{table*}

\begin{table*}
    \centering
    \caption{Five examples for the image recognition task with confidence levels. Please refer to Section \ref{sup:dataset} for the question prompts.}
    \begin{tabular}{p{2cm}p{4cm}p{4cm}p{4cm}}
        \hline
        \textbf{Question} & \textbf{GPT-4V} & \textbf{Gemini Pro Vision} & \textbf{Correct} \\
        \textbf{Number} \\
         \hline

        36 & Mount Fuji (90\%) & Mount Fuji (99\%) & Mount Fuji\\ 

        \multicolumn{3}{c}{} \\

        12 & 90 Japanese gravestones and memorial tablets (95\%) & A graveyard (80\%) & Japanese graveyard/cemetery\\ 

        \multicolumn{3}{c}{} \\

        29 & Japanese cuisine, including sushi, sashimi, and tempura (confidence 95\%) & Japanese food (100\%) &  Japanese food or cuisine, also acceptable that it is food model, called Shokuhin Sampuru in Japanese\\

        \multicolumn{3}{c}{} \\

        22 & Tokyo (80\%) & Tokyo (80\%) & Japan, Tokyo, Shinjuku Gyoen National Garden\\ 

        \multicolumn{3}{c}{} \\

        26b & Tokyo, Japan (70\%) & Osaka Castle (80\%) & Castle Osaka, Osaka, Japan\\ \hline
        
    \end{tabular}
    
    \vspace*{33em}
\end{table*}

\end{document}